\PassOptionsToPackage{numbers,sort&compress}{natbib}
\documentclass{article}

\usepackage[preprint]{neurips_2026}

\usepackage[utf8]{inputenc}
\usepackage[T1]{fontenc}
\usepackage{amsmath,amssymb,amsthm}
\usepackage{mathtools}
\usepackage{hyperref}
\usepackage{url}
\usepackage{booktabs}
\usepackage{multirow}
\usepackage{amsfonts}
\usepackage{subcaption}
\usepackage{amsmath}
\usepackage{amssymb}
\usepackage{nicefrac}
\usepackage{microtype}
\usepackage{xcolor}
\definecolor{impgreen}{HTML}{2E8B57}
\usepackage{wrapfig}
\usepackage{graphicx}
\usepackage{enumitem}
\usepackage{algorithm}
\usepackage{algpseudocode}
\usepackage{float}
\usepackage[most]{tcolorbox}

\usepackage{titlesec}

\newtcblisting{templatebox}{
    breakable,
    enhanced,
    colback=blue!2,
    colframe=blue!55!black,
    boxrule=0.7pt,
    arc=1.8mm,
    left=1mm,
    right=1mm,
    top=1mm,
    bottom=1mm,
    listing only,
    listing options={basicstyle=\ttfamily\small,breaklines=true,columns=fullflexible,keepspaces=true}
}

\newtheorem{remark}{Remark}

\title{Ranking-Aware Calibration for Reliable Multimodal Reinforcement Learning}

\author{
\textsuperscript{1}Peng Cui\thanks{Equal contribution.},
~~\textsuperscript{2}Boyao Yang\footnotemark[1],
~~\textsuperscript{1}Jun Zhu\thanks{Corresponding author.} \\
\textsuperscript{1} Dept. of Comp. Sci. \& Tech., Institute for AI, BNRist Center, \\
Tsinghua-Bosch Joint ML Center, THBI Lab, Tsinghua University, Beijing 100084, China \\
\textsuperscript{2} Dept. of Automation, Tsinghua University, Beijing 100084, China \\
\texttt{xpeng.cui@gmail.com, boyaoyang16@gmail.com, dcszj@tsinghua.edu.cn}
}

\begin{document}
\addtocontents{toc}{\protect\setcounter{tocdepth}{-1}}
\maketitle

\begin{abstract}
Reinforcement learning (RL) post-training has substantially improved the reasoning accuracy of vision-language models, yet the resulting policies remain poorly calibrated. Terminal correctness rewards provide no gradient that penalizes confident errors more than uncertain ones, and no signal that ties confidence to the quality of the supporting visual evidence. This mismatch becomes especially severe under corrupted or ambiguous visual inputs, where models continue to report high confidence on wrong answers.
We introduce \textbf{R}anking-\textbf{A}ware \textbf{C}alibration (RAC), a training-time framework that supervises confidence using two comparison signals that group-based RL already produces at no additional labeling cost. The \emph{ranking-aware group loss} enforces that a better rollout receives higher confidence than a worse one within the same prompt. The \emph{clean--corrupted pairwise loss} enforces that confidence attenuates as visual evidence degrades. Because the ranking signal forces the policy to distinguish between correct and incorrect reasoning paths, it also reinforces task accuracy beyond what correctness rewards alone produce. Both losses require no external confidence annotations and integrate naturally with group-based RL post-training.
We instantiate RAC on Qwen2.5-VL and InternVL-3.5 backbones and evaluate on six multimodal reasoning benchmarks under clean and corrupted inputs. Empirical results show that the ranking-aware loss substantially improves task accuracy by teaching the policy to discriminate between better and worse reasoning, while the pairwise corruption loss reduces calibration error under degraded inputs. Their combination achieves the best calibration across all tested backbones while significantly improving accuracy in the majority of settings. Our code is available at \url{https://github.com/Stellaris167/RAC}.
\end{abstract}

\section{Introduction}
The integration of visual perception and language reasoning has driven rapid progress in artificial intelligence.
Modern vision-language models (VLMs) and multimodal large language models (MLLMs) can now solve tasks ranging from spatial understanding to multi-step mathematical reasoning~\citep{radford2021clip,alayrac2022flamingo,li2023blip2,liu2023visualinstruction,bai2023qwenvl,wang2024qwen2vl,bai2025qwen25vl,zhu2025internvl3}.
A major driver of these gains has been reinforcement learning (RL) with verifiable outcome rewards: by optimizing directly for correctness, models learn to produce longer reasoning traces, revise intermediate conclusions, and self-correct, yielding substantial performance improvements on visual mathematics and science benchmarks~\citep{shao2024deepseekmath,deepseek2025r1,wang2025mpo}.

However, better reasoning accuracy does not automatically translate into better
calibration. This problem has since been studied extensively in text-only LLMs through several distinct directions, including post-hoc recalibration of model probabilities via temperature scaling and adaptive variants~\citep{kadavath2022know,tian2023just,xie2024ats}, verbalized confidence elicitation~\citep{xiong2024llmuncertainty,zhang2024calibrating}, and training-time interventions such as representation editing~\citep{ji2025verbaluncertainty},
listener-aware fine-tuning~\citep{stengel2024lacie}, calibration-aware reward shaping during RL post-training~\citep{damani2025rlcr}, and analyses of how chain-of-thought behavior itself affects calibration~\citep{yoon2025reasoningcalibration}.
By contrast, calibration in multimodal reasoning models remains comparatively underexplored, and recent evidence suggests that confidence can become especially fragile when visual evidence is degraded or ambiguous~\citep{saxena2026vlmrobustbench,sui2025benchc}.
Prior work has shown that calibration often degrades under distribution shift and input corruption~\citep{ovadia2019trust,hendrycks2019benchmarking,minderer2021revisiting}.
An MLLM may inspect a corrupted image, generate a fluent rationale, arrive at an incorrect answer, and still report high confidence. The underlying issue is structural: terminal correctness rewards distinguish right answers from wrong ones, but they do not explicitly penalize \emph{confident errors} more than uncertain ones, nor do they require confidence to track the quality of the supporting visual evidence.

This gap is especially damaging because it directly amplifies hallucination. When a model reasons confidently from weak or corrupted visual input, the failure is not merely an isolated prediction error but a form of authoritative misreasoning that can propagate to downstream decision-making systems~\citep{li2023pope,chen2024multiobjecthallucination,dang2026instinct}.
Recent analyses show that VLMs frequently assign high verbalized confidence to hallucinated answers, and that the gap between stated certainty and actual correctness widens sharply under visual corruption~\citep{dang2026instinct,byun2026medicalvqa}.
In systems that use confidence for abstention, reranking, or query routing, this failure mode is particularly costly. Existing methods fail to address the core challenge in multimodal reasoning: confidence should respond to the quality of visual evidence. It should fall when evidence quality falls, and among multiple candidate solutions to the same question, better reasoning should receive higher confidence than worse reasoning. These directional properties are absent from text-only objectives and cannot be retroactively imposed by post-hoc rescaling of a frozen model. Moreover, learning to rank confidence across rollouts of the same prompt also provides a harder discrimination signal that directly reinforces correct reasoning, yielding accuracy gains that correctness rewards alone do not produce.

We therefore treat calibration as a native objective within multimodal RL. Our key observation is that group-based RL already provides two forms of supervision at no additional labeling cost. For each prompt, the policy samples multiple rollouts with known correctness, yielding a \emph{within-prompt} comparison axis. Across prompts, any clean input can be paired with a synthetically corrupted version of the same question, yielding an \emph{across-evidence} comparison axis. These two axes capture exactly the directional properties missing from standard outcome rewards, without requiring any external confidence annotations. We instantiate this idea in the \textbf{RAC} (\textbf{R}anking-\textbf{A}ware \textbf{C}alibration) framework, which encodes the two axes as margin-based losses integrated directly into the RL reward. The \emph{ranking-aware group loss} enforces that, within the same prompt, a better rollout receives higher confidence than a worse one. The \emph{clean—corrupted pairwise loss} enforces that confidence decreases when the same question is presented with weaker visual evidence, with the required gap scaling with corruption severity. Together, these objectives teach the model to align confidence with both reasoning quality and evidence quality during RL training itself, and as a byproduct of learning to rank confidence across rollouts, the policy is also pushed toward more accurate reasoning.

Our contributions are threefold:
\begin{itemize}[leftmargin=*, itemsep=2pt, topsep=2pt]
    \item \textbf{A relative supervision view of RL calibration.}\enspace We identify two directional properties that standard outcome-based multimodal RL fails to enforce, and show that both can be supervised from comparisons already produced by RL sampling, namely within-prompt rollout groups and matched clean--corrupted pairs, without requiring external confidence labels.
    \item \textbf{The RAC framework.}\enspace We formalize these two supervision axes as margin-based ranking and pairwise losses and distribute them back into sequence-level rewards, yielding a unified calibration objective that integrates directly into any group-based RL post-training pipeline without requiring changes to the underlying optimizer or model architecture.
    \item \textbf{Empirical evaluation under visual corruption.}\enspace We instantiate RAC on Qwen2.5-VL-3B/7B-Instruct and InternVL-3.5-8B, and evaluate it on six multimodal reasoning benchmarks under clean and progressively corrupted inputs, covering eight ImageNet-C operators and five severity levels. Across all backbones, the ranking-aware loss yields substantial accuracy gains and the pairwise corruption loss yields substantial calibration improvements, with their combination achieving the best performance on both dimensions in most settings.
\end{itemize}

\section{Method}
\label{sec:method}
This section presents the RAC framework in three parts.
We first review the outcome-based multimodal RL setting and the group-based sampling geometry on which our method relies (Section~\ref{sec:rl_baselines}). We then introduce the two RAC calibration objectives: one for ordering confidence within a prompt, and one for enforcing confidence attenuation across clean and corrupted views (Section~\ref{sec:calibration_obj}). Finally, we explain how these objectives are converted into shaped rewards and optimized in a unified training loop (Section~\ref{sec:optimization}).

\begin{figure*}[t]
\vspace{-10pt}
    \centering
    \includegraphics[width=0.90\textwidth]{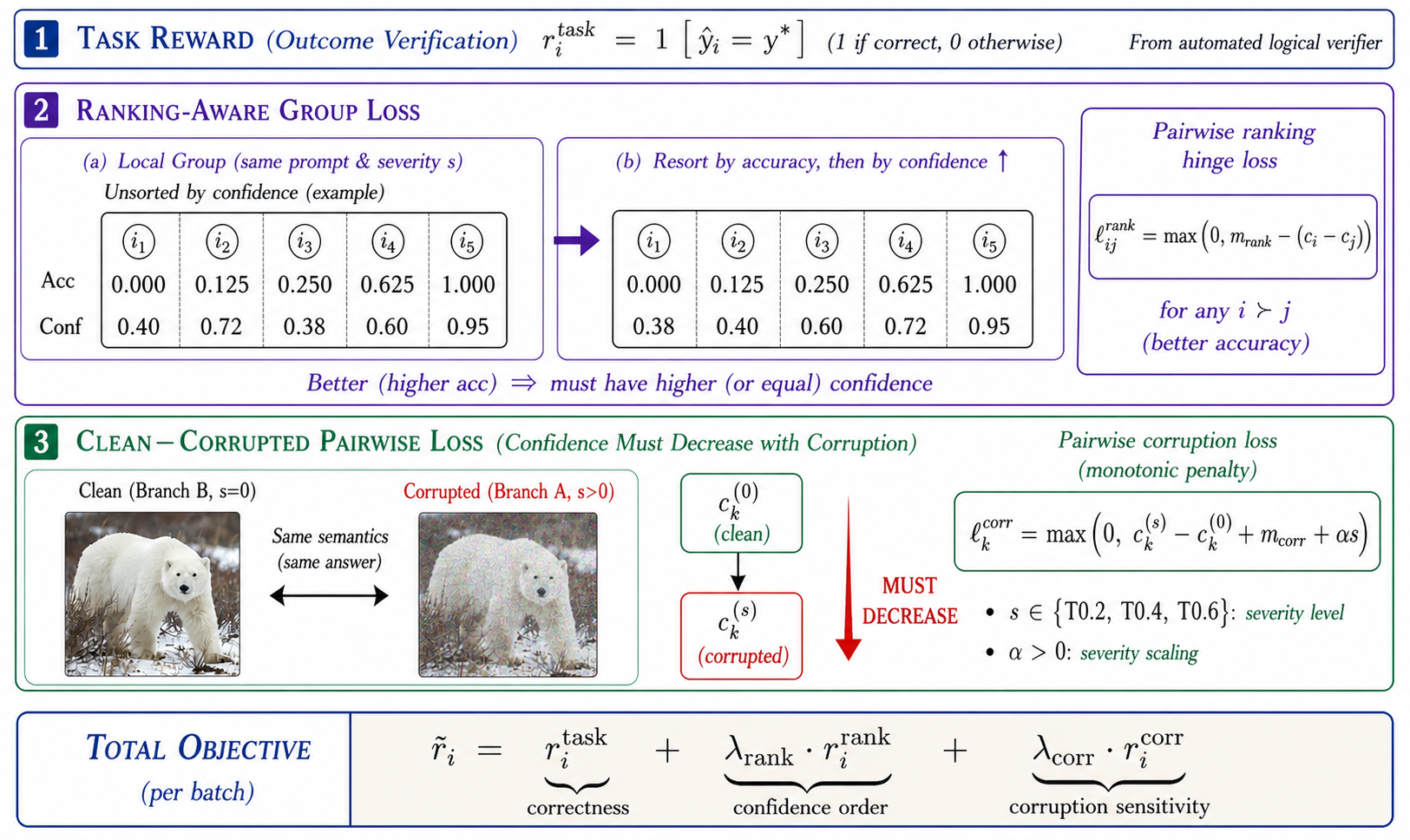}
    \vspace{-6pt}
    \caption{\textbf{Overview of the RAC Calibration Architecture.} The framework is structured around three reinforcement signals: (1) the standard outcome verification task reward, (2) the Ranking-Aware Group Loss, which asserts that higher confidence should coordinate with greater answer correctness within the same rollout group, and (3) the Clean--Corrupted Pairwise Loss, which enforces directional confidence attenuation when the visual input is synthetically degraded.}
    \label{fig:rac_framework}
\vspace{-8pt}
\end{figure*}

\subsection{Outcome-Based Reinforcement Learning}
\label{sec:rl_baselines}
In outcome-based multimodal RL, the goal is to fine-tune a pre-trained policy $\pi_\theta$ to maximize verifier-derived rewards. Classical alignment methods often relied on Proximal Policy Optimization (PPO), which stabilizes policy updates using a learned value baseline $V_\phi$. However, training a full multimodal critic that mirrors the policy can introduce substantial memory overhead for modern VLMs.

Recent sample-efficient RL methods avoid the explicit critic by using multiple rollouts from the same prompt to construct relative baselines~\citep{shao2024deepseekmath,ahmadian2024backtobasics}. This paradigm has since been extended to multimodal reasoning, with recent work adapting group-based RL to train vision-language models on verifiable visual tasks~\citep{wang2025scs,huang2025vision}. Let $x \in \mathcal{D}$ denote a multimodal prompt containing text and image input.
For each prompt, the policy samples a group of $n$ rollouts $\mathcal{G}(x)=\{y_1,\dots,y_n\}$ and receives task rewards $r_1,\dots,r_n$ from a verifier.
This prompt-level rollout group is the basic structural unit on which RAC operates.

Group Relative Policy Optimization (GRPO)~\citep{shao2024deepseekmath} directly estimates the baseline by standardizing the returns within the group $\mathcal{G}(x)$. The advantage for a rollout $y_i$ is computed as:
\begin{equation}
\label{eq:grpo}
    A_i^{\text{GRPO}}
    = \frac{r_i - \mu_r}{\sigma_r}, \qquad
    \mu_r = \frac{1}{n}\sum_{j=1}^{n} r_j, \qquad
    \sigma_r = \sqrt{\frac{1}{n}\sum_{j=1}^{n}(r_j - \mu_r)^2}.
\end{equation}
GRPO supports a standard clipped policy optimization objective with a
reference model regularizer:
{\small \begin{equation}
\label{eq:surrogate}
    \mathcal{J}_{\mathrm{RL}}(\theta)
    =
    \mathbb{E}_{x \sim \mathcal{D},\; \{y_i\}_{i=1}^{n} \sim \pi_{\mathrm{old}}}
    \left[
    \frac{1}{n}\sum_{i=1}^{n}
    \min\!\left(
        \rho_i(\theta) A_i,\;
        \mathrm{clip}\!\bigl(\rho_i(\theta), 1-\epsilon, 1+\epsilon\bigr) A_i
    \right)
    \right]
    - \beta_{\mathrm{KL}}\,\mathbb{D}_{\mathrm{KL}}(\pi_\theta \| \pi_{\mathrm{ref}}),
\end{equation}}
where $\rho_i(\theta)=\frac{\pi_\theta(y_i \mid x)}{\pi_{\mathrm{old}}(y_i \mid x)}$ is the importance ratio, $\epsilon$ controls the clipping range, and $\beta_{\mathrm{KL}}$ weights deviation from the reference model. Because GRPO already organizes optimization around prompt-level sets of alternative trajectories, it provides exactly the comparison structure needed to supervise confidence \emph{relatively} rather than absolutely. This structure is precisely what RAC exploits: rollout groups provide the within-prompt comparisons used by the ranking loss, while clean--corrupted pairs provide the across-evidence comparisons used by the corruption loss.

\subsection{Ranking-aware Calibration Objectives}
\label{sec:calibration_obj}
\textbf{Confidence-bearing rollouts.}
RAC assumes that each sampled rollout contains not only a rationale and final answer, but also an explicit scalar confidence value. Concretely, for a prompt $x$, one rollout sampled from $\pi_\theta$ produces a completion $y_i$, a parsed confidence $c_i \in [0,1]$, a verifier-based correctness signal $a_i \in \{0,1\}$, and a format indicator $f_i \in \{0,1\}$ that checks
whether the response follows the required output schema. We therefore represent one rollout as $(y_i,c_i,a_i,f_i)$ and sample a prompt-level group
\begin{equation}
\label{eq:rollout_group}
\mathcal{G}(x)=\{(y_i,c_i,a_i,f_i)\}_{i=1}^{n},
\qquad (y_i,c_i) \sim \pi_\theta(\cdot \mid x).
\end{equation}
Within a single prompt, question difficulty is largely controlled. As a result, comparing rollouts from the same question provides a clean local signal for confidence quality: if several candidate solutions compete for one prompt, the better solution should be assigned higher confidence than the worse one. This relative supervision is more trustworthy than forcing each confidence value toward an absolute probability target, since no calibrated ground-truth target is available in outcome-reward RL.

\textbf{Ranking-aware group calibration.}
For each rollout group $\mathcal{G}(x)$ (Eq.~\ref{eq:rollout_group}), we define the valid comparison set
\begin{equation}
\label{eq:comparison_set}
\mathcal{Q}(x)=\{(i,j): a_i > a_j,\; f_i=1,\; f_j=1\}.
\end{equation}
Each pair compares a better rollout against a worse rollout from the same prompt.
We then impose a margin-based ranking loss
\begin{equation}
\label{eq:rank_loss}
\ell^{\mathrm{rank}}_{ij}=\max\bigl(0,\; m_{\mathrm{rank}}-(c_i-c_j)\bigr),
\end{equation}
where $m_{\mathrm{rank}}>0$ is a confidence margin. The loss vanishes only when the better rollout receives sufficiently higher confidence. Aggregating over all valid pairs (Eq.~\ref{eq:rank_loss}) yields
\begin{equation}
\label{eq:rank_agg}
L_{\mathrm{rank}}(x)=
\frac{1}{|\mathcal{Q}(x)|}
\sum_{(i,j)\in\mathcal{Q}(x)} \ell^{\mathrm{rank}}_{ij},
\end{equation}
with the convention that $L_{\mathrm{rank}}(x)=0$ when no valid comparison exists.In practice, this groupwise loss is converted into rollout-level shaping terms by accumulating, for each sampled rollout, the ranking violations of all valid pairs in which it participates.

\begin{remark}[Why relative, not absolute?]
Forcing confidence toward absolute targets (e.g., $c_i \to a_i$) is appealing but suffers from a mismatch: the binary correctness label $a_i$ is not itself a well-calibrated probability target, especially for borderline or partially correct responses. In contrast, the pairwise margin only requires that better reasoning attract more certainty than worse reasoning \emph{within the same context}. This relative objective is both weaker and better aligned with the information available from online RL rollouts. Furthermore, enforcing relative ordering within a prompt group provides a stronger training signal than a correctness reward alone: the policy must not only produce a correct answer but also assign it distinguishably higher confidence than incorrect alternatives, which directly reinforces the quality of the underlying reasoning.
\end{remark}


\textbf{Clean--corrupted pairwise calibration.}
The ranking term governs confidence \emph{within} a prompt, but it does not say how confidence should change when the quality of the underlying evidence changes. We therefore introduce paired clean and corrupted views $x^{(0)}$ and $x^{(s)}$ of the same question, where $s \in [0,1]$ denotes corruption severity and $s=0$ is the clean input. The desired behavior is directional: if two inputs share the same semantic content but one contains weaker visual evidence, the corrupted view should not retain the same level of confidence as the clean one.

For each clean prompt $x^{(0)}$ and its corrupted counterpart $x^{(s)}$, we sample two rollout groups and form slot-wise pairs
\begin{equation}
\label{eq:slot_pair}
\bigl((y_k^{(0)},c_k^{(0)},a_k^{(0)}),(y_k^{(s)},c_k^{(s)},a_k^{(s)})\bigr).
\end{equation}
Here, the index $k$ identifies corresponding draws from the two groups inside the same training iteration. It is used to organize the pairwise supervision and does not assume token-level trajectory correspondence between clean and corrupted rollouts. We then impose a directional margin that favors higher confidence on the clean view:
\begin{equation}
\label{eq:corr_loss}
\ell^{\mathrm{corr}}_{k}=
\max\bigl(0,\; c_k^{(s)}-c_k^{(0)}+m_{\mathrm{corr}}+\alpha s\bigr),
\end{equation}
where $m_{\mathrm{corr}}>0$ is a base margin and $\alpha \ge 0$ scales the desired confidence drop with corruption severity. This objective does not require every corrupted sample to be low-confidence in isolation. Rather, it expresses a cleaner comparative rule: holding semantic content fixed, the corrupted view should not be assigned the same or higher confidence than the clean view without incurring a penalty.

Aggregating over the slot-wise pairs (Eq.~\ref{eq:corr_loss}) yields:
\begin{equation}
\label{eq:corr_agg}
L_{\mathrm{corr}}(x^{(0)},x^{(s)})=
\frac{1}{n}\sum_{k=1}^{n} \ell^{\mathrm{corr}}_{k}.
\end{equation}

\textbf{Connection to contrastive learning.}
The corruption loss is structurally similar in spirit to supervised contrastive objectives \citep{khosla2020supcon}, which use relative relationships rather than absolute labels. Here, however, the comparison is imposed directly on the scalar confidence rather than on a representation vector: the clean and corrupted views of the same question form a matched pair, and the clean view is required to be \emph{more confident}. This makes it possible to supervise calibration without any explicit confidence annotation.

\subsection{Optimization and Training Procedure}
\label{sec:optimization}
\begin{algorithm}[t]
\caption{Ranking-Aware Multimodal RL}
\label{alg:calibration_rl}
\begin{algorithmic}[1]
\Require Multimodal policy $\pi_\theta$, clean prompt $x^{(0)}$, corruption operator $T_s$, severity distribution $p(s)$, rollout count $n$.
\State Sample corruption severity $s \sim p(s)$.
\State Form corrupted input $x^{(s)} = T_s(x^{(0)})$.
\For{$k = 1$ \textbf{to} $n$}
    \State Sample clean rollout $(y_k^{(0)}, c_k^{(0)}) \sim \pi_\theta(\cdot \mid x^{(0)})$.
    \State Sample corrupted rollout $(y_k^{(s)}, c_k^{(s)}) \sim \pi_\theta(\cdot \mid x^{(s)})$.
    \State Obtain verifier scores $a_k^{(0)}, a_k^{(s)} \in \{0, 1\}$.
    \State Obtain format indicators $f_k^{(0)}, f_k^{(s)} \in \{0, 1\}$.
\EndFor
\State Find valid clean comparisons $\mathcal{Q}(x^{(0)}) = \{(i,j): a_i^{(0)} > a_j^{(0)}, f_i^{(0)} = f_j^{(0)} = 1\}$.
\State Find valid corrupted comparisons $\mathcal{Q}(x^{(s)})$ analogously.
\State Evaluate ranking group losses $L_{\mathrm{rank}}(x^{(0)})$ and $L_{\mathrm{rank}}(x^{(s)})$ (Eq.~\ref{eq:rank_agg}).
\State Evaluate pairwise corruption loss $L_{\mathrm{corr}}(x^{(0)}, x^{(s)})$ (Eq.~\ref{eq:corr_agg}).
\State Compute shaped sequence rewards $\tilde{r}_k^{(0)}$ and $\tilde{r}_k^{(s)}$ (Eq.~\ref{eq:shaped_reward}) using task returns and calibration terms.
\State Update policy $\pi_\theta$ using GRPO over the combined mini-batch.
\end{algorithmic}
\end{algorithm}

\textbf{Unified objective.}
The two losses play complementary roles. The ranking term specifies how confidence should be ordered among alternative solutions to the same prompt, while the corruption term specifies how confidence should change when evidence quality degrades. Their combined calibration objective is
\begin{equation}
\label{eq:cal_objective}
\mathcal{L}_{\mathrm{cal}}=
\mathbb{E}_{x \sim \mathcal{D}}\bigl[L_{\mathrm{rank}}(x)\bigr]
+ \beta\, \mathbb{E}_{(x^{(0)},x^{(s)}) \sim \mathcal{D}_{\mathrm{pair}}}
\bigl[L_{\mathrm{corr}}(x^{(0)},x^{(s)})\bigr],
\end{equation}
where $\mathcal{D}$ is the prompt distribution, $\mathcal{D}_{\mathrm{pair}}$ is the paired clean--corrupted distribution, and $\beta$ balances the two relative supervision signals. In practice, we optimize a sequence-level RL objective whose shaped reward is
\begin{equation}
\label{eq:shaped_reward}
\tilde r_i = r^{\mathrm{task}}_i
+ \lambda_{\mathrm{rank}} r^{\mathrm{rank}}_i
+ \lambda_{\mathrm{corr}} r^{\mathrm{corr}}_i
+ \lambda_{\mathrm{fmt}} r^{\mathrm{fmt}}_i,
\end{equation}
with the outer objective
\begin{equation}
\label{eq:outer_objective}
\mathcal{J}(\theta)=\mathbb{E}_{\pi_\theta}\left[\frac{1}{n}\sum_{i=1}^{n} \tilde r_i\right].
\end{equation}
Here, $r^{\mathrm{task}}_i$ is the benchmark-specific correctness reward, $r^{\mathrm{rank}}_i$ and $r^{\mathrm{corr}}_i$ are rollout-level shaping terms induced by the two RAC objectives (Eqs.~\ref{eq:rank_agg} and~\ref{eq:corr_agg}), and $r^{\mathrm{fmt}}_i$ is a lightweight regularizer that stabilizes the structured output format and can be decayed during training. The outer policy update is carried out with GRPO. RAC therefore leaves the underlying RL optimizer unchanged, and it only changes the reward signal so that confidence behavior becomes part of the learning objective.

\section{Experiment}
This section presents our experimental setup, main results, and ablation studies. We evaluate RAC across six multimodal backbones, six reasoning benchmarks, and five corruption severity levels to assess whether training-time confidence supervision improves calibration without sacrificing accuracy. We further ablate the effect of training corruption severity on calibration generalization and conduct a sensitivity analysis on the loss hyperparameters to validate key design choices.
\subsection{Experimental Setup}
\textbf{Backbones.} 
We evaluate three open multimodal checkpoints that span a representative range of model scales and architectures: Qwen2.5-VL-3B-Instruct and Qwen2.5-VL-7B-Instruct \citep{bai2025qwen25vl}, and InternVL-3.5-8B \citep{zhu2025internvl3}. These models are strong enough to expose nontrivial confidence failures while remaining tractable for RL post-training. More details are provided in Appendix~\ref{app:models_datasets}.

\textbf{Hyperparameters.} 
All experiments are implemented using the VERL framework~\citep{sheng2024hybridflow}. We sample $n = 8$ rollouts per prompt with a global batch size of 512 sequences. We use a structured chat template that elicits a verbalized scalar confidence alongside the reasoning chain and final answer (Appendix~\ref{app:chat_template}). Responses are decoded with a sampling temperature of $0.6$, a top-$p$ threshold of $0.95$, and a top-$k$ cutoff of $20$, with a maximum response length of 3{,}072 tokens. The policy is optimized with a learning rate of $1\times10^{-6}$ under cosine decay. The ranking margin is set to $m_{\mathrm{rank}} = 0.05$, and reward weights in Eq.~\ref{eq:shaped_reward} are $\lambda_{\mathrm{rank}} = 0.2$ and $\lambda_{\mathrm{corr}} = 0.3$. The format weight $\lambda_{\mathrm{fmt}}$ is initialized at $1.0$ and linearly decayed to $0.3$ to prevent format supervision from competing with calibration signals at convergence.

\textbf{Training mixture.}
The training set is formed by merging appropriate subsets of M3CoT~\citep{chen2024m3cot}, Geometry3K~\citep{lu2021geometry3k}, and ScienceQA~\citep{lu2022scienceqa}, retaining only multiple-choice items. Each training pair consists of a corrupted branch $A$ and a clean branch $B$. Branch $A$ is sampled from a mixed-corruption stream in which 50\% of samples are drawn at severity T0.2, 40\% at T0.4, and 10\% at T0.6, while branch $B$ remains clean. This severity mixture is designed to provide meaningful directional contrast between branches while preserving sufficient visual content for the model to reason about. Image corruptions follow the ImageNet-C protocol~\citep{hendrycks2019benchmarking}, covering 8 diverse noise types (detailed in Appendix~\ref{sec:corruptions}), with the specific operator chosen uniformly at random for each sample.

\paragraph{Evaluation setup.} We evaluate on six multiple-choice benchmarks spanning mathematics, science, and broader multi-discipline domains: M3CoT~\citep{chen2024m3cot}, ScienceQA~\citep{lu2022scienceqa}, MathVerse~\citep{zhang2024mathverse}, MMMU~\citep{yue2024mmmu}, We-Math~\citep{wang2024wemath}, and MathVision~\citep{wang2024mathvision}, retaining only multiple-choice items to enable unambiguous answer verification and reliable confidence elicitation via a fixed response schema. Corruption severity $T \in \{0.0, 0.2, 0.4, 0.6, 0.8, 1.0\}$ ranges from clean ($\mathrm{T0.0}$) to maximally degraded ($\mathrm{T1.0}$), with the specific corruption operator drawn uniformly at random from the 8 types, yielding $6 \times 6 = 36$ evaluation subsets in total. We report three metrics macro-averaged across all six benchmarks: Accuracy, Expected Calibration Error (ECE)~\citep{naeini2015obtaining}, and Brier score~\citep{brier1950forecast}. Formal definitions of ECE and Brier score are provided in Appendix~\ref{sec:eval_metrics}. Results are further aggregated into three severity bands: \textbf{Clean} ($\mathrm{T0.0}$), \textbf{Mild} ($\mathrm{T0.2}$--$\mathrm{T0.4}$ average), and \textbf{Severe} ($\mathrm{T0.6}$--$\mathrm{T1.0}$ average).

\subsection{Main Results}
Table~\ref{tab:main_results} reveals a consistent and interpretable decomposition across all backbone families. The pairwise corruption loss (\textit{+Pair}) substantially reduces ECE and Brier score relative to Vanilla-RL while leaving task accuracy largely unchanged, consistent with its role in teaching the policy to attenuate confidence as visual evidence weakens rather than directly improving answer correctness. The ranking-aware loss (\textit{+Rank}), in contrast, consistently raises task accuracy by encouraging the policy to assign higher confidence to correct reasoning paths, though the resulting confidence increase also produces a modest rise in ECE. Combining both objectives under \textit{+Pair+Rank} achieves the best calibration across nearly all settings while exceeding Vanilla-RL accuracy, confirming that the two losses are complementary rather than conflicting: the ranking loss strengthens reasoning discrimination, while the corruption loss restores confidence reliability under
degraded evidence.

\begin{table*}[t]
\vspace{-6pt}
    \centering
    \small
    \setlength{\tabcolsep}{4pt}
    \caption{Macro-averaged metrics across 6 benchmarks grouped by noise severity: \textbf{Clean} ($\mathrm{T0.0}$), \textbf{Mild} ($\mathrm{T0.2}$--$\mathrm{T0.4}$ average), and \textbf{Severe} ($\mathrm{T0.6}$--$\mathrm{T1.0}$ average). Vanilla-RL denotes models post-trained with GRPO. \textbf{Bold} = best Avg per model; \underline{underline} = second best. {\color{impgreen}$\uparrow$\!/$\downarrow$} in \textit{+Pair+Rank} rows show relative gain over Base.}
    \label{tab:main_results}
    \resizebox{\textwidth}{!}{%
    \begin{tabular}{lcccccccccccc}
        \specialrule{1.5pt}{0pt}{3pt}
        \multirow{2}{*}{Variant} & \multicolumn{4}{c}{Accuracy $\uparrow$} & \multicolumn{4}{c}{ECE $\downarrow$} & \multicolumn{4}{c}{Brier $\downarrow$} \\
        \cmidrule(lr){2-5} \cmidrule(lr){6-9} \cmidrule(lr){10-13}
        & Clean & Mild & Severe & Avg & Clean & Mild & Severe & Avg & Clean & Mild & Severe & Avg \\
        \specialrule{1.5pt}{3pt}{5pt}
        \multicolumn{13}{c}{\textbf{InternVL-3.5-8B}} \\[3pt]
        \midrule
        Vanilla-RL       & 0.437 & 0.441 & 0.436 & 0.438 & 0.354 & 0.359 & 0.364 & 0.359 & 0.384 & 0.388 & 0.388 & 0.386 \\
        +Pair      & 0.529 & 0.521 & 0.509 & 0.520 & 0.350 & 0.356 & 0.369 & 0.358 & 0.350 & 0.355 & 0.368 & 0.358 \\
        +Rank      & 0.566 & 0.560 & 0.545 & \textbf{0.557} & 0.320 & 0.337 & 0.350 & \underline{0.336} & 0.322 & 0.338 & 0.349 & \underline{0.336} \\
        +Pair+Rank & 0.552 & 0.546 & 0.533 & \underline{0.544}~{\color{impgreen}\tiny$\uparrow$24.2\%} & 0.303 & 0.309 & 0.320 & \textbf{0.311}~{\color{impgreen}\tiny$\downarrow$13.4\%} & 0.307 & 0.315 & 0.325 & \textbf{0.316}~{\color{impgreen}\tiny$\downarrow$18.1\%} \\
        \specialrule{0.8pt}{2pt}{5pt}
        \multicolumn{13}{c}{\textbf{Qwen2.5-VL-3B-Instruct}} \\[3pt]
        \midrule
        Vanilla-RL       & 0.527 & 0.516 & 0.500 & 0.514 & 0.319 & 0.328 & 0.342 & 0.329 & 0.341 & 0.348 & 0.359 & 0.349 \\
        +Pair      & 0.519 & 0.511 & 0.495 & 0.508 & 0.295 & 0.304 & 0.317 & \underline{0.305} & 0.316 & 0.326 & 0.335 & \underline{0.326} \\
        +Rank      & 0.557 & 0.546 & 0.530 & \underline{0.544} & 0.329 & 0.338 & 0.352 & 0.340 & 0.351 & 0.358 & 0.369 & 0.359 \\
        +Pair+Rank & 0.590 & 0.582 & 0.568 & \textbf{0.580}~{\color{impgreen}\tiny$\uparrow$12.8\%} & 0.287 & 0.294 & 0.302 & \textbf{0.294}~{\color{impgreen}\tiny$\downarrow$10.6\%} & 0.308 & 0.315 & 0.319 & \textbf{0.314}~{\color{impgreen}\tiny$\downarrow$10.0\%} \\
        \specialrule{0.8pt}{2pt}{5pt}
        \multicolumn{13}{c}{\textbf{Qwen2.5-VL-7B-Instruct}} \\[3pt]
        \midrule
        Vanilla-RL       & 0.630 & 0.631 & 0.618 & 0.626 & 0.323 & 0.326 & 0.340 & 0.330 & 0.335 & 0.339 & 0.352 & 0.342 \\
        +Pair      & 0.623 & 0.625 & 0.612 & 0.620 & 0.299 & 0.303 & 0.314 & \underline{0.305} & 0.312 & 0.316 & 0.328 & \underline{0.319} \\
        +Rank      & 0.660 & 0.661 & 0.648 & \textbf{0.656} & 0.333 & 0.336 & 0.350 & 0.340 & 0.345 & 0.349 & 0.362 & 0.352 \\
        +Pair+Rank & 0.643 & 0.634 & 0.624 & \underline{0.634}~{\color{impgreen}\tiny$\uparrow$1.3\%} & 0.270 & 0.287 & 0.294 & \textbf{0.284}~{\color{impgreen}\tiny$\downarrow$13.9\%} & 0.285 & 0.302 & 0.313 & \textbf{0.300}~{\color{impgreen}\tiny$\downarrow$12.3\%} \\
        \specialrule{0.8pt}{2pt}{-5pt}
    \end{tabular}\vspace{-6pt}%
    }
\vspace{-6pt}
\end{table*}
\begin{figure}[t]
\centering
\includegraphics[width=0.80\linewidth]{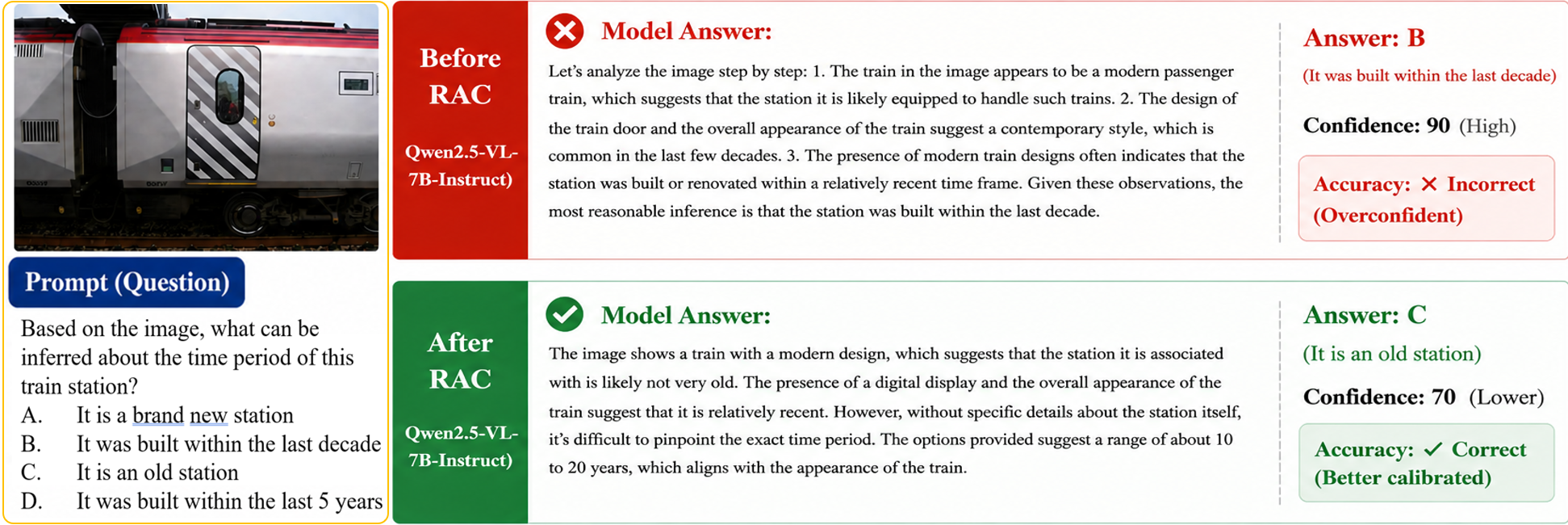}
\caption{Qualitative case study on Qwen2.5-VL-7B-Instruct. Before RAC, the model produces an incorrect answer (B) with high confidence (90), reflecting overconfident reasoning from ambiguous visual evidence. After RAC training, the same model arrives at the correct answer (C) with appropriately reduced confidence (70), demonstrating improved alignment between stated certainty and evidence quality.}
\label{fig:case_study}
\vspace{-10pt}
\end{figure}

The gains are most pronounced for InternVL-3.5-8B, which enters training with below-average calibration and a relatively disordered confidence signal. The \textit{+Rank} variant alone lifts average accuracy from 0.438 to 0.557 (an absolute gain of 11.9\%, the largest single-component accuracy improvement across all experiments), suggesting that ranking supervision provides a particularly strong corrective signal for models whose initial confidence ordering is weakly aligned with answer quality. Applying the full \textit{+Pair+Rank} framework then reduces ECE from 0.359 to 0.311 and Brier from 0.386 to 0.316, the largest absolute calibration improvement in the table. A nuance arises in Qwen2.5-VL-7B, where \textit{+Rank} alone achieves the highest accuracy (0.656) while \textit{+Pair+Rank} recovers the best calibration (ECE 0.284) at a slight accuracy cost (0.634). This mild tension reflects the pairwise loss encouraging lower confidence on degraded inputs, which can marginally attenuate the accuracy-boosting effect of the ranking term at larger model scale. The net result remains a Pareto improvement over Vanilla-RL on both dimensions, confirming that RAC generalizes across model families without architectural assumptions. Figure~\ref{fig:case_study} provides a qualitative illustration of this behavior. On the same ambiguous visual question, the model assigns high confidence (90\%) to an incorrect answer before RAC training, but after training produces the correct answer with appropriately reduced confidence (70\%), reflecting the improved alignment between stated certainty and evidence quality that the two calibration losses are designed to instill.

\begin{figure}[t]
\vspace{-6pt}
\centering
\begin{subfigure}[b]{0.49\textwidth}
\centering
\includegraphics[width=\linewidth]{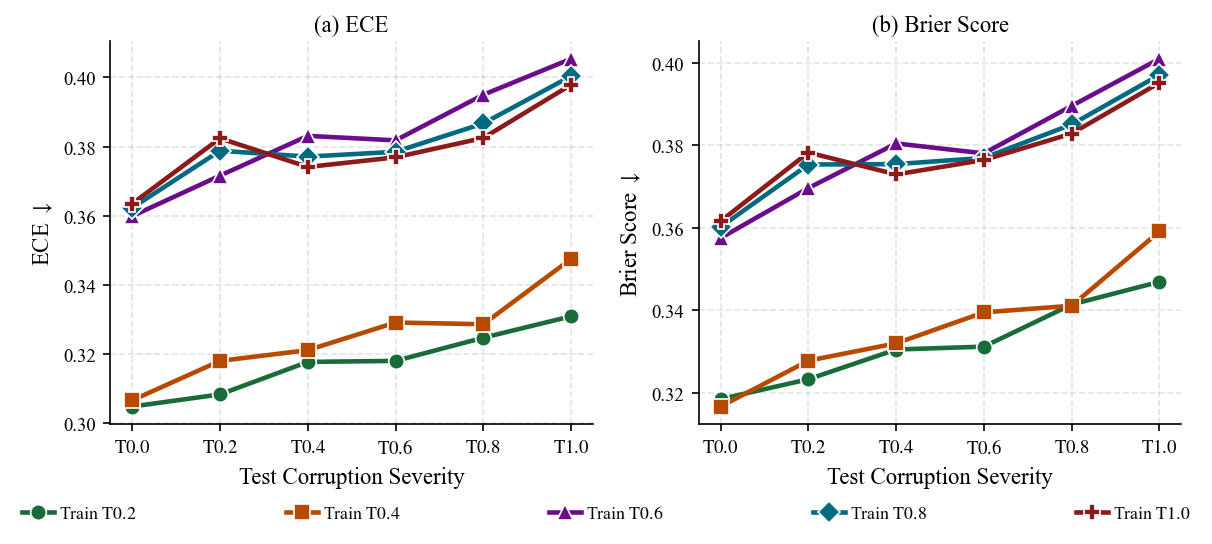}
\caption{Severity-transfer curves.}
\label{fig:ablation_severity}
\end{subfigure}
\hfill
\begin{subfigure}[b]{0.49\textwidth}
\centering
\includegraphics[width=\linewidth]{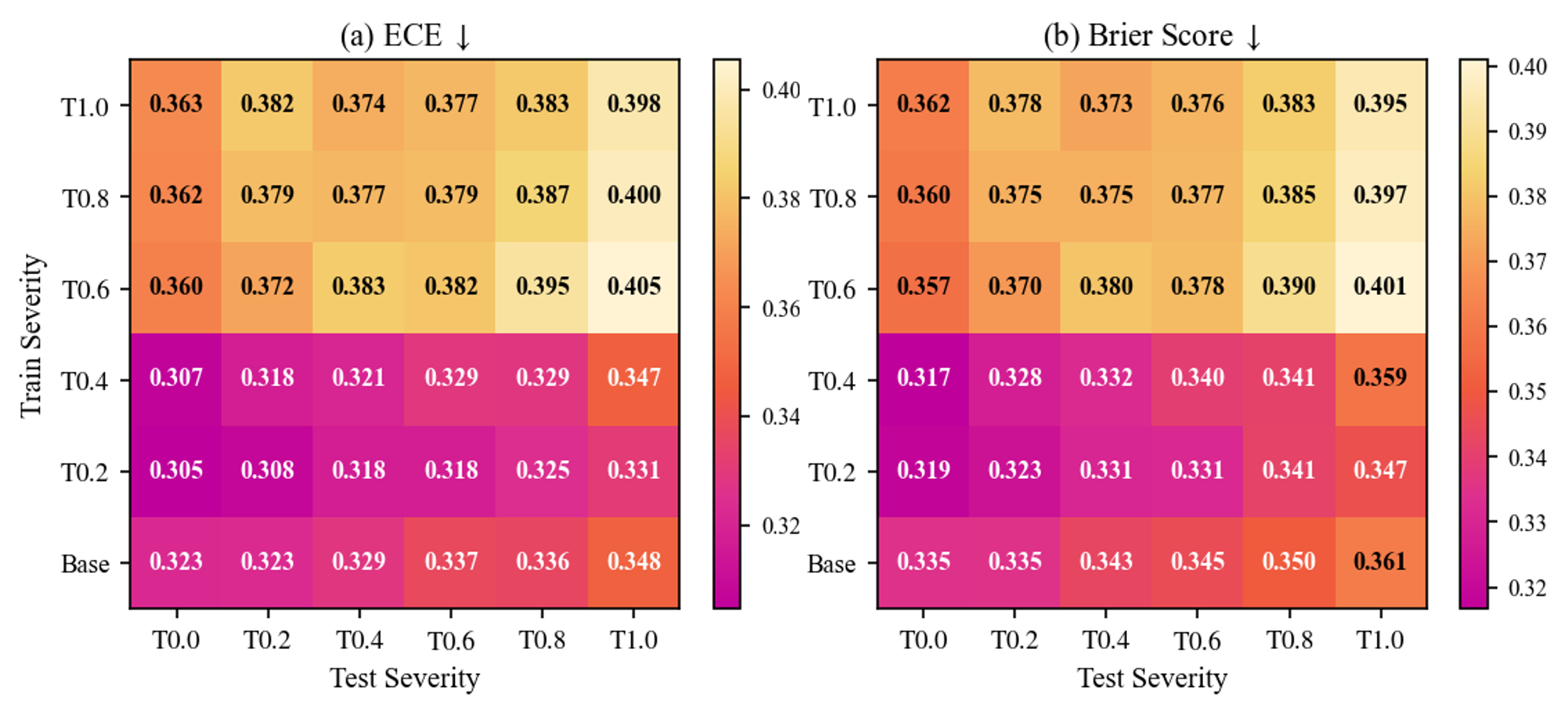}
\caption{Severity-transfer heatmaps.}
\label{fig:ablation_heatmap}
\end{subfigure}
\caption{\textbf{Effect of training corruption severity on calibration transfer} for Qwen2.5-VL-7B-Instruct. Models trained at mild corruption levels ($\mathrm{T0.2}$, $\mathrm{T0.4}$) maintain consistently lower ECE and Brier scores across all test severities, while those trained at $\mathrm{T0.6}$ and above exhibit uniformly degraded calibration. The severity-transfer curves in (a) separate into two distinct clusters at $\mathrm{T0.6}$, a pattern confirmed by the heatmaps in (b).}
\label{fig:ablation_combined}
\vspace{-9pt}
\end{figure}

\begin{figure}[t]
\vspace{-5pt}
\centering
\begin{subfigure}[b]{0.45\linewidth}
    \centering
    \includegraphics[width=\linewidth]{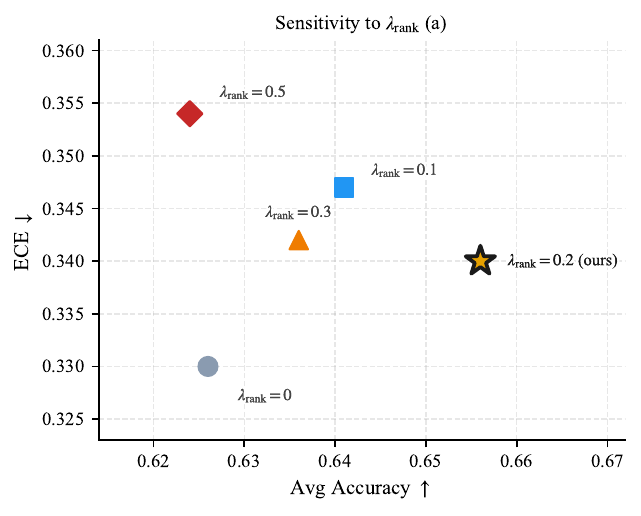}
    \caption{$\lambda_{\mathrm{rank}}$ sweep ($+$Rank).}
    \label{fig:lambda_sensitivity_a}
\end{subfigure}\hfill
\begin{subfigure}[b]{0.45\linewidth}
    \centering
    \includegraphics[width=\linewidth]{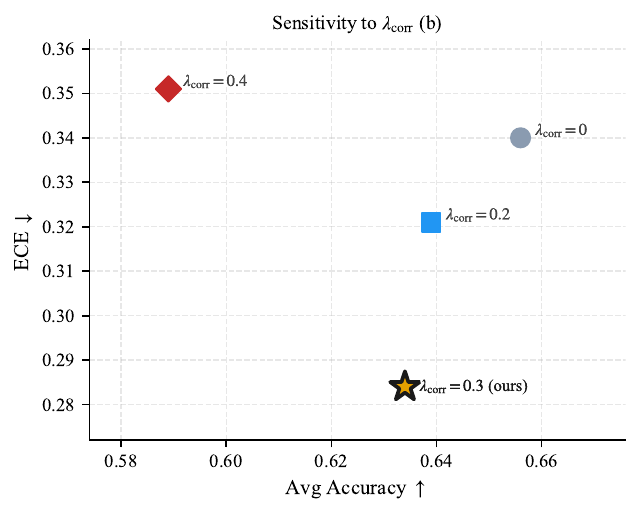}
    \caption{$\lambda_{\mathrm{corr}}$ sweep ($+$Rank$+$Pair, fixed $\lambda_{\mathrm{rank}}=0.2$).}
    \label{fig:lambda_sensitivity_b}
\end{subfigure}
\caption{Hyperparameter sweeps on Qwen2.5-VL-7B-Instruct. Each point shows macro-averaged Accuracy (x-axis, higher is better) and ECE (y-axis, lower is better). (a) Increasing $\lambda_{\mathrm{rank}}$ peaks accuracy at $0.2$ with controlled ECE. (b) Increasing $\lambda_{\mathrm{corr}}$ reduces ECE up to $0.3$ (selected), but $0.4$ degrades accuracy and increases ECE.}
\label{fig:lambda_sensitivity}
\vspace{-10pt}
\end{figure}

\textbf{Effect of training corruption severity on calibration generalization.}
To study how training-time corruption strength affects calibration transfer, we train five variants of Qwen2.5-VL-7B-Instruct using different \emph{fixed} corruption levels for branch $A$: $\mathrm{T0.2}$, $\mathrm{T0.4}$, $\mathrm{T0.6}$, $\mathrm{T0.8}$, and $\mathrm{T1.0}$. Branch $B$ remains clean throughout, and all other hyperparameters are held constant. Each variant is then evaluated on all six test severities, reporting macro-averaged Accuracy, ECE, and Brier score across the six benchmarks.
Figures~\ref{fig:ablation_severity} and~\ref{fig:ablation_heatmap} reveal a clear non-monotonic relationship between training corruption strength and calibration quality. Introducing a mild corruption signal consistently benefits the model: the $\mathrm{T0.2}$ variant achieves lower ECE across all test severities and retains the highest task accuracy among all fixed-corruption configurations. Beyond this moderate level, however, further increasing training corruption proves detrimental. The $\mathrm{T0.6}$, $\mathrm{T0.8}$, and $\mathrm{T1.0}$ variants all exhibit uniformly degraded calibration and weaker accuracy, indicating that excessively heavy corruption destabilizes the supervision signal rather than improving robustness. These results suggest that an effective training corruption level strikes a balance: strong enough to expose the model to degraded visual evidence and encourage appropriate confidence attenuation, yet moderate enough to preserve sufficient visual content and avoid inducing systematically under-confident predictions.

\subsection{Ablation Study}
\textbf{Hyperparameter sensitivity.}
We evaluate the sensitivity of both reward weights by sweeping each over a range of values while holding all other hyperparameters fixed, using Qwen2.5-VL-7B-Instruct throughout. For $\lambda_{\mathrm{rank}}$, we train five \textit{+Rank} variants with $\lambda_{\mathrm{rank}} \in \{0, 0.1, 0.2, 0.3, 0.5\}$, where $\lambda_{\mathrm{rank}} = 0$ reduces to Vanilla RL. For $\lambda_{\mathrm{corr}}$, we train four \textit{+Rank+Pair} variants with $\lambda_{\mathrm{corr}} \in \{0, 0.2, 0.3, 0.4\}$, where $\lambda_{\mathrm{corr}} = 0$ reduces to \textit{+Rank} alone. Figure~\ref{fig:lambda_sensitivity} plots macro-averaged accuracy against ECE for each configuration, with the lower-right region indicating better joint performance. For $\lambda_{\mathrm{rank}}$ (Figure \ref{fig:lambda_sensitivity_a}), the setting $\lambda_{\mathrm{rank}} = 0.2$ achieves the highest accuracy ($0.656$) with ECE remaining well-controlled at $0.340$. Larger values degrade both metrics, suggesting that an excessively strong ranking penalty impairs rather than reinforces the confidence-ordering objective. For $\lambda_{\mathrm{corr}}$ (Figure \ref{fig:lambda_sensitivity_b}), increasing the weight from $0$ to $0.3$ reduces ECE from $0.340$ to $0.284$ at a modest accuracy cost, identifying $\lambda_{\mathrm{corr}} = 0.3$ as the optimal setting. At $\lambda_{\mathrm{corr}} = 0.4$, accuracy drops sharply to $0.589$ and ECE rises again, indicating that an excessively large corruption penalty destabilizes training. Together, these sweeps confirm that $\lambda_{\mathrm{rank}} = 0.2$ and $\lambda_{\mathrm{corr}} = 0.3$ consistently identify the best accuracy-calibration trade-off across both reward dimensions.

\section{Related Work}

\textbf{Reinforcement learning for alignment.}
Reinforcement learning has increasingly become the standard for post-training large language and multimodal models. Early human-feedback alignment relied on Proximal Policy Optimization (PPO)~\citep{schulman2017ppo,ouyang2022instructgpt} to match human preferences, while DPO~\citep{rafailov2023dpo} streamlined this by directly optimizing from offline preference pairs. As the field shifted toward eliciting complex reasoning, the alignment objective evolved from mimicking human style to optimizing verifiable correctness. Outcome-based RL has thus gained prominence for reasoning tasks~\citep{shao2024deepseekmath,deepseek2025r1,wang2025mpo}, training models against rule-based verifiers to develop long chain-of-thought logic and autonomous self-correction. In multimodal models, this reasoning search is even more demanding, as the policy must additionally align linguistic logic with visual perception. To reduce the memory overhead of critic networks, GRPO estimates baselines directly from multiple rollouts of the same prompt~\citep{shao2024deepseekmath} and has since been adapted for visual reasoning tasks~\citep{wang2025scs,liu2025visual,huang2025vision}.
\vspace{5pt}

\textbf{Confidence calibration and visual shift.}
Calibration research examines whether a model's stated confidence reflects its true probability of being correct~\citep{niculescu2005predicting,guo2017calibration,brier1950forecast}. A well-established finding is that calibration degrades sharply under distribution shift and common corruptions~\citep{ovadia2019trust,minderer2021revisiting}. Within generative models, the evaluation paradigm has evolved from softmax logits to verbalized confidence and semantic uncertainty estimates~\citep{desai2020calibration,kadavath2022know,kuhn2023semantic}. For multimodal models, the failure mode is more severe: visual anomalies cascade into hallucinated reasoning, whereby models assign high expressed certainty to incorrect answers generated from degraded inputs~\citep{li2023pope,chen2024multiobjecthallucination,dang2026instinct,byun2026medicalvqa}. Post-hoc methods such as temperature scaling and Bayesian binning offer partial relief~\citep{guo2017calibration,naeini2015obtaining}, and recent work specifically targets VLM calibration via decoupled rescaling~\citep{xiao2026vlcalibration}. However, all such approaches operate on a frozen model and cannot alter what the policy has learned to be certain about.

\vspace{5pt}
\textbf{Calibration and uncertainty in LLMs.}
A growing body of work examines whether large language models can reliably express their own uncertainty. On the post-hoc side, temperature-based recalibration of RLHF-tuned model probabilities~\citep{kadavath2022know,tian2023just,xie2024ats} and verbalized confidence elicitation~\citep{xiong2024llmuncertainty,zhang2024calibrating} have been shown to partially recover calibration that post-training degrades, though all such approaches operate on a frozen model. Training-time interventions go further by making calibration a direct learning signal, through representation editing~\citep{ji2025verbaluncertainty}, listener-aware fine-tuning~\citep{stengel2024lacie}, and calibration rewards integrated into RL post-training~\citep{damani2025rlcr}. Analyses of emergent behavior further suggest that reasoning-oriented RL can improve confidence expression as a byproduct, as models engaging in self-verification and backtracking tend to assign confidence more accurately~\citep{yoon2025reasoningcalibration}. These studies collectively demonstrate that uncertainty can be elicited, calibrated, and trained in text-only LLMs, but they do not address the multimodal setting in which confidence must additionally respond to the quality of visual evidence.

\section{Conclusion}
We presented RAC, a training-time calibration framework that addresses a structural gap in multimodal RL post-training: existing outcome-reward objectives optimize for answer correctness but provide no signal governing how confidence should track reasoning quality or visual evidence quality. RAC closes this gap with two relative supervision signals arising naturally from group-based RL sampling. The ranking-aware group loss enforces that correct reasoning attracts higher confidence than incorrect reasoning within the same prompt, improving task accuracy as a byproduct. The clean--corrupted pairwise loss enforces confidence attenuation as visual evidence weakens, reducing ECE and Brier score. Together, their combination achieves improvements of up to 18.1\% in Brier and 13.9\% in ECE relative to Vanilla RL, with both objectives requiring no external annotations and no changes to the underlying optimizer.


Several directions remain open for future work. Extending RAC to adaptive corruption curricula that progressively diversify severity and operator type could further strengthen calibration transfer to unseen degradations. Furthermore, one limitation is that RAC is currently evaluated on multiple-choice benchmarks, where binary correctness labels provide unambiguous reward signals for confidence calibration. So a longer-term direction is applying RAC to open-ended generation settings, where binary correctness labels are unavailable and confidence must instead be grounded in preference or semantic similarity signals.

\bibliographystyle{unsrtnat}
\bibliography{neurips_2026}

\clearpage
\appendix
\renewcommand{\contentsname}{Appendices}
\addtocontents{toc}{\protect\setcounter{tocdepth}{2}}
\tableofcontents
\clearpage

\section{Details of Models and Datasets}
\label{app:models_datasets}

\paragraph{Models.}
We fine-tune three open-source vision-language models.
\textbf{Qwen2.5-VL-3B-Instruct} is a 3-billion-parameter instruction-tuned model from the Qwen2.5-VL family~\citep{bai2025qwen25vl}, released under the Apache~2.0 License (\url{https://huggingface.co/Qwen/Qwen2.5-VL-3B-Instruct}).
\textbf{Qwen2.5-VL-7B-Instruct} is the 7-billion-parameter counterpart~\citep{bai2025qwen25vl}, also released under the Apache~2.0 License (\url{https://huggingface.co/Qwen/Qwen2.5-VL-7B-Instruct}).
\textbf{InternVL2.5-8B} is an 8-billion-parameter model from the InternVL3 family~\citep{zhu2025internvl3}, released under the MIT License (\url{https://huggingface.co/OpenGVLab/InternVL2_5-8B}).
These three checkpoints were selected to cover a representative range of scales and architectures while remaining tractable for RL post-training.

\paragraph{Training datasets.}
\textbf{M$^3$CoT}~\citep{chen2024m3cot} is a multi-domain, multi-step, multi-modal chain-of-thought benchmark covering science, mathematics, and commonsense reasoning (MIT License; \url{https://huggingface.co/datasets/LightChen233/M3CoT}).
\textbf{Geometry3K}~\citep{lu2021geometry3k} is a formal geometry problem set with diagrams and structured annotations (MIT License; \url{https://huggingface.co/datasets/hiyouga/geometry3k}).
\textbf{ScienceQA}~\citep{lu2022scienceqa} is a multimodal multiple-choice benchmark spanning natural science, language science, and social science topics (MIT License; \url{https://huggingface.co/datasets/derek-thomas/ScienceQA}).
We retain only the multiple-choice items from each source in the training mixture.

\paragraph{Evaluation datasets.}
\textbf{M$^3$CoT}~\citep{chen2024m3cot} is also used for evaluation on its held-out test split (MIT License; see above for the dataset link).
\textbf{ScienceQA}~\citep{lu2022scienceqa} is also used for evaluation on its held-out test split (MIT License; see above for the dataset link).
\textbf{MathVerse}~\citep{zhang2024mathverse} tests whether models can truly perceive diagrams in visual math problems (MIT License; \url{https://huggingface.co/datasets/AI4Math/MathVerse}).
\textbf{MMMU}~\citep{yue2024mmmu} is a massive multi-discipline multimodal benchmark spanning 30 university subjects and 183 subfields (Apache~2.0; \url{https://huggingface.co/datasets/MMMU/MMMU}).
\textbf{We-Math}~\citep{wang2024wemath} provides hierarchical visual math problems designed to probe knowledge acquisition and generalization (CC~BY-NC~4.0; \url{https://huggingface.co/datasets/We-Math/We-Math}).
\textbf{MathVision}~\citep{wang2024mathvision} is a competition-level mathematical problem set with visual contexts spanning 16 disciplines (CC~BY-NC-SA~4.0; \url{https://huggingface.co/datasets/mathvision-bench/MathVision}).

\section{Chat Template}
\label{app:chat_template}

\begin{templatebox}
"\n Please reason step by step and follow this exact response schema:\n"
"<think>your reasoning</think>\n"
"<answer>your final answer</answer>\n"
"<confidence>your confidence</confidence>\n"
"If options are provided, put only the single option letter inside <answer>.\n"
"Otherwise, put only the final value or short expression inside <answer>.\n"
"Inside <confidence>, output exactly one decimal number between 0.0 and 1.0.\n"
"Start your response with <think> and do not output any text before <think> or after </confidence>."
\end{templatebox}

\section{Image Corruptions and Examples}
\label{sec:corruptions}

\paragraph{Corruption algorithms.} We implement 8 visual degradation operators following the ImageNet-C robustness protocol \citep{hendrycks2019benchmarking}, each mapped across 5 discrete severity levels ($\mathrm{T0.2}$ to $\mathrm{T1.0}$). These corruptions are summarized in Table~\ref{tab:corruptions}.

Importantly, during both training pair construction and evaluation, the specific corruption operator applied to any given image is sampled uniformly at random from these 8 types. Two images at the same severity level (e.g., both at $\mathrm{T0.4}$) may thus be degraded by entirely different operators (e.g., Fog vs.\ Elastic Transform), preventing the policy from overfitting to any single noise pattern and encouraging robustness to diverse visual degradation.

\begin{table}[ht]
    \centering
    \small
    \caption{Summary of visual degradation operators and their corresponding severity parameter scales ($\mathrm{T0.2}$ to $\mathrm{T1.0}$).}
    \label{tab:corruptions}
    \begin{tabular}{llp{4cm}}
        \toprule
        \textbf{Corruption Type} & \textbf{Description} & \textbf{Severity Scale ($\mathrm{T0.2}$ -- $\mathrm{T1.0}$)} \\
        \midrule
        \textbf{Gaussian Noise} & Additive zero-mean Gaussian noise. & Std.\ dev.\ $0.08$ \textrightarrow $0.38$ (rel.\ to $255$). \\
        \textbf{Shot Noise} & Poisson noise mimicking low-light sensor defects. & Scaled photon counts $60$ \textrightarrow $3$. \\
        \textbf{Impulse Noise} & Salt-and-pepper noise affecting random pixels. & Random fraction $3\%$ \textrightarrow $27\%$. \\
        \textbf{Fog} & Additive synthetic diamond-square plasma fractal. & Increasing fog thickness/intensity. \\
        \textbf{Brightness} & HSV space Value-channel manual adjustments. & Uniform offset $0.1$ \textrightarrow $0.5$. \\
        \textbf{Contrast} & Blending toward the image's mean pixel intensity. & Decreasing standard deviation. \\
        \textbf{Elastic Transform} & Localized warping via affine transformations. & Increasing displacement field vectors. \\
        \textbf{JPEG Compression} & Heavy reduction in image storage quality. & \texttt{libjpeg} quality $25$ \textrightarrow $7$. \\
        \bottomrule
    \end{tabular}
\end{table}

\paragraph{Dataset degradation ladder.}

\begin{figure*}[ht]
    \centering
    \includegraphics[width=\textwidth]{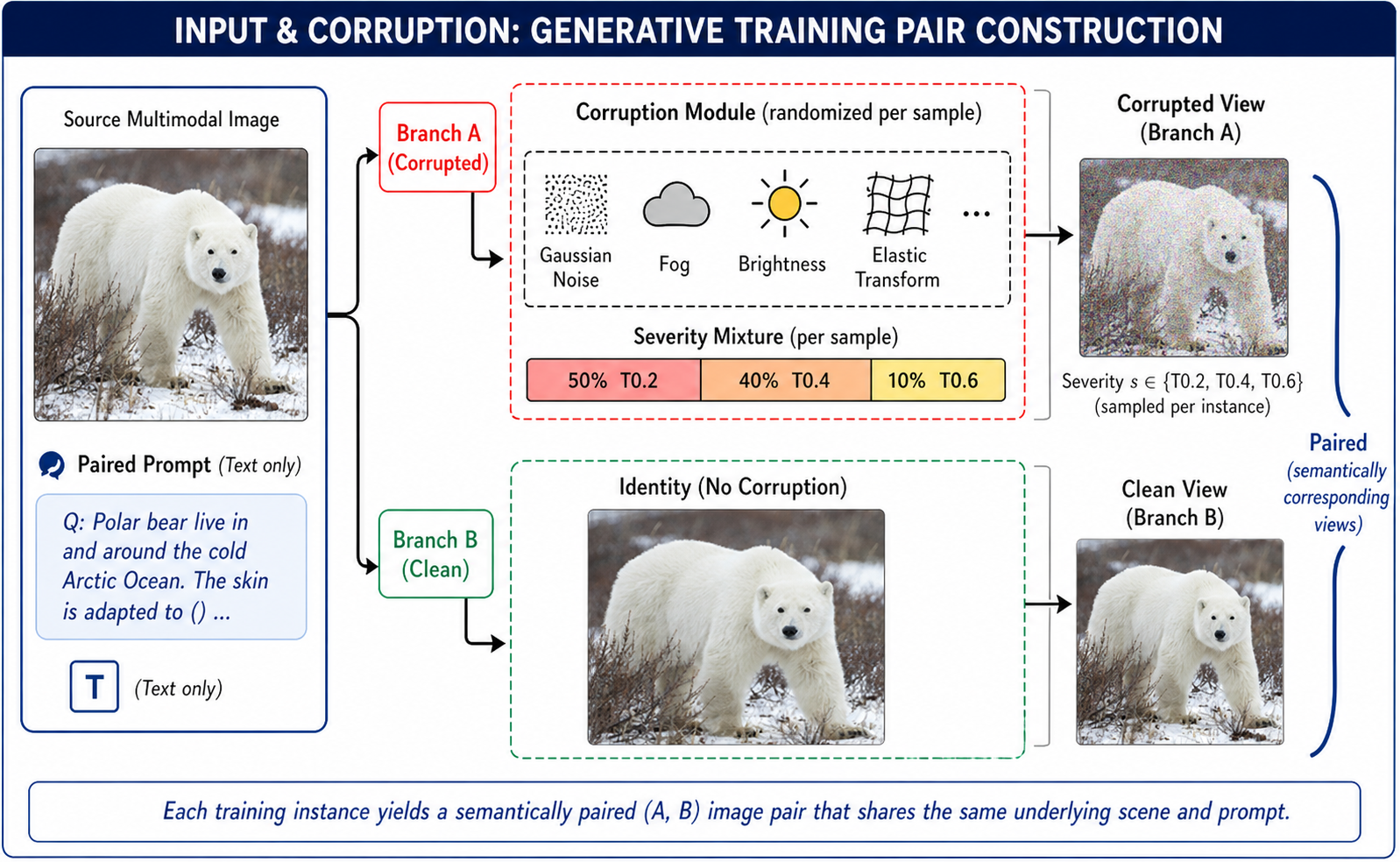}
    \caption{\textbf{Generative Training Pair Construction.} The matched multimodal branches underpinning our method: \textit{Branch A (Corrupted)} applies randomly sampled visual perturbation (following the ImageNet-C protocol), juxtaposed against the untouched source in \textit{Branch B (Clean)}. This paired structure provides the clean--corrupted comparisons used in the RAC pairwise loss (Figure~\ref{fig:rac_framework}).}
    \label{fig:pair_construction}
\end{figure*}

\begin{figure}[htbp]
    \centering
    \setlength{\tabcolsep}{3pt}
    \renewcommand{\arraystretch}{0.5}
    \begin{tabular}{ccc}
        \multicolumn{3}{c}{\textbf{M3CoT Degradation Ladder}} \\[2pt]
        \textbf{Clean} & $\mathbf{T0.2}$ & $\mathbf{T0.4}$ \\
        \includegraphics[width=0.31\linewidth]{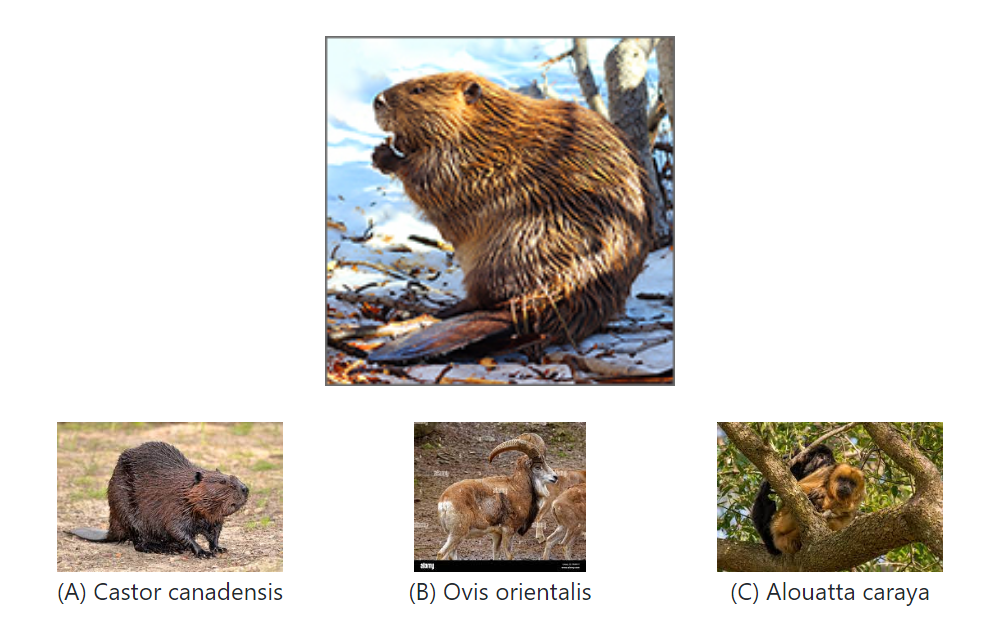} & \includegraphics[width=0.31\linewidth]{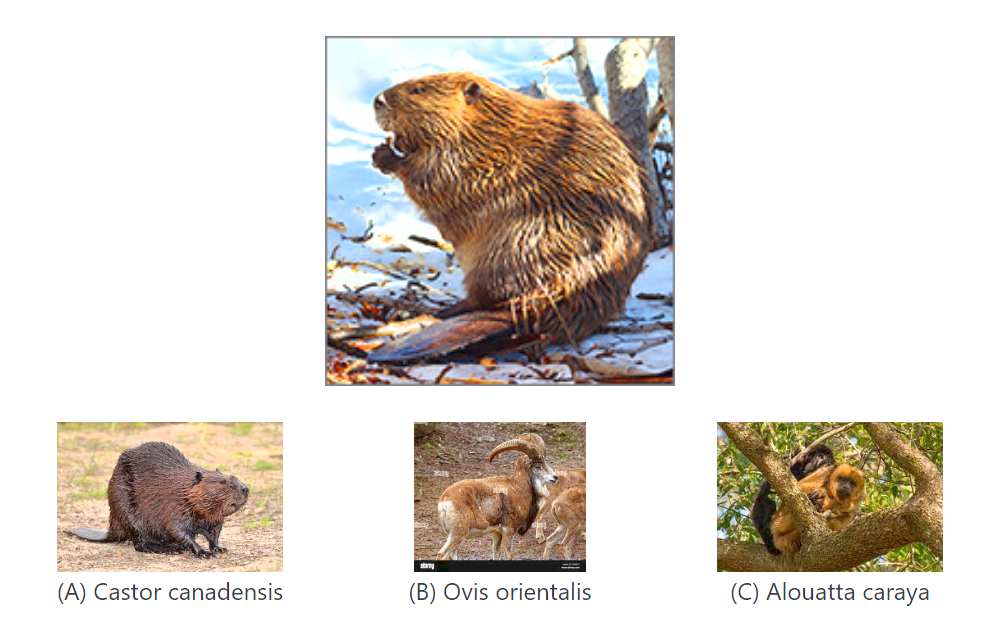} & \includegraphics[width=0.31\linewidth]{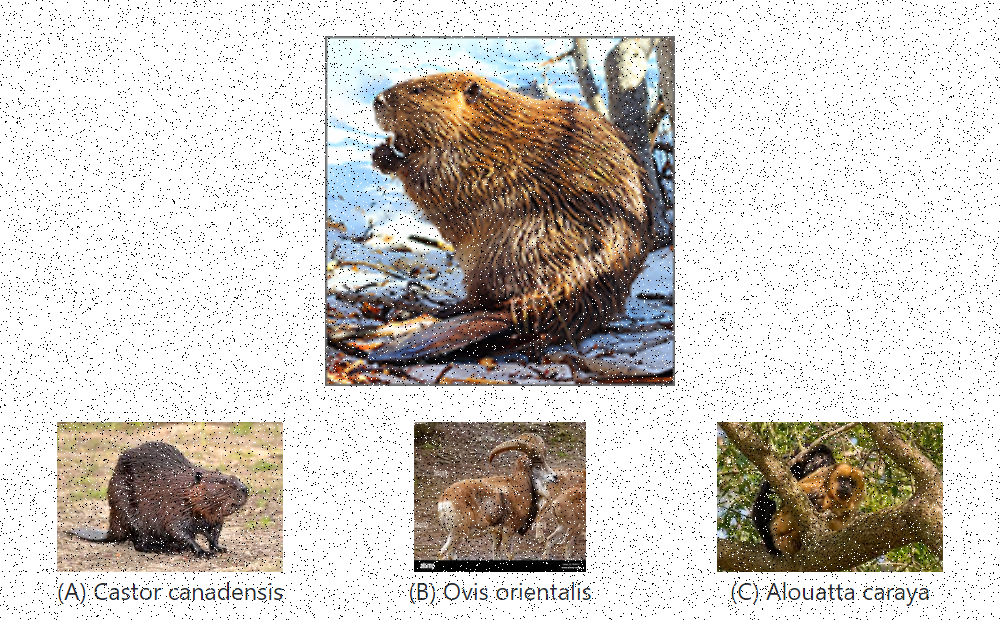} \\[4pt]
        $\mathbf{T0.6}$ & $\mathbf{T0.8}$ & $\mathbf{T1.0}$ \\
        \includegraphics[width=0.31\linewidth]{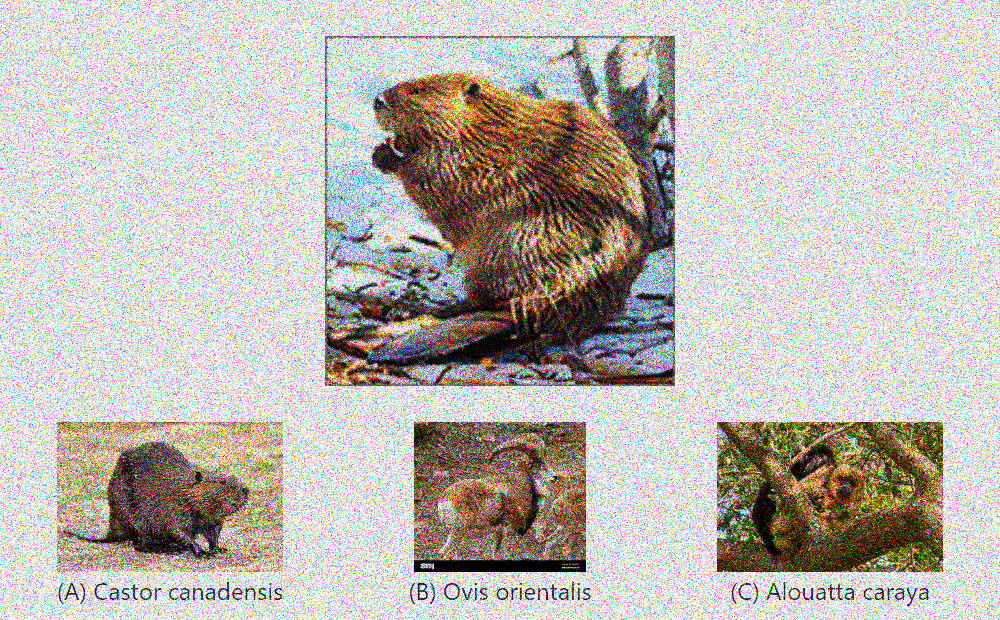} & \includegraphics[width=0.31\linewidth]{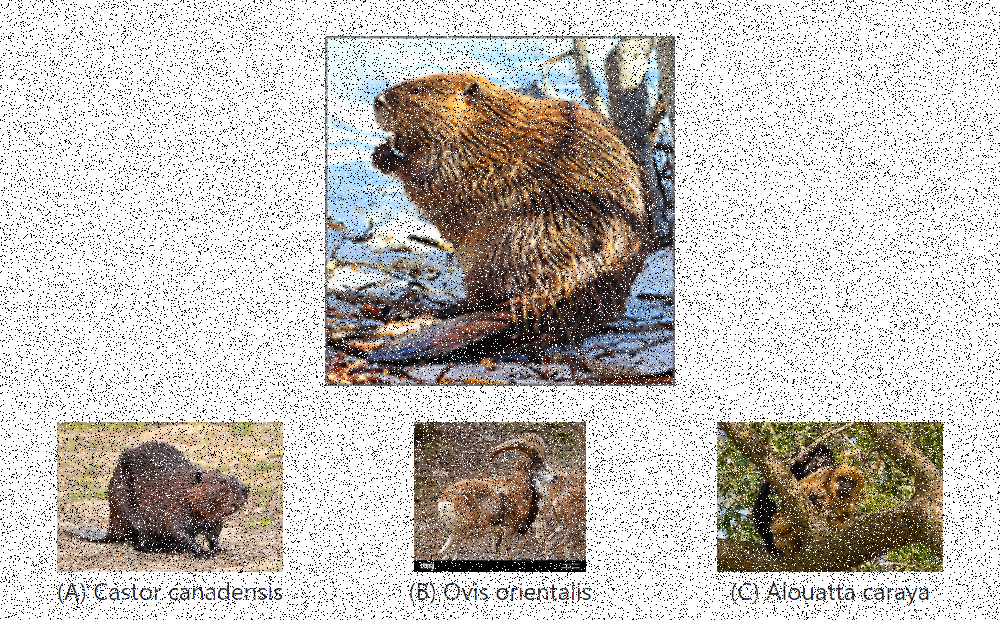} & \includegraphics[width=0.31\linewidth]{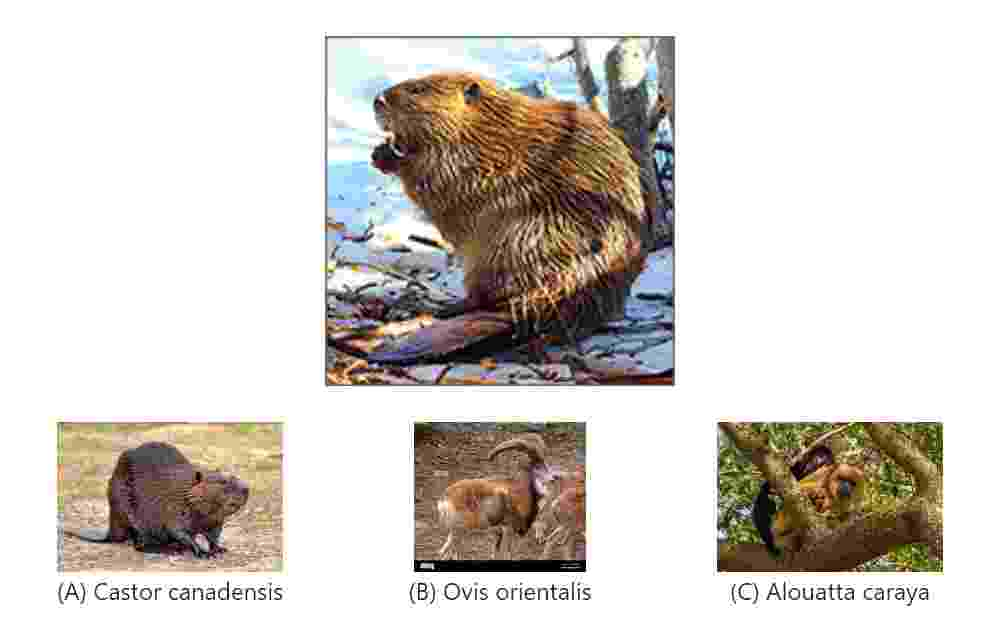} \\[8pt]
        \midrule
        \\[-2pt]
        \multicolumn{3}{c}{\textbf{ScienceQA Degradation Ladder}} \\[2pt]
        \textbf{Clean} & $\mathbf{T0.2}$ & $\mathbf{T0.4}$ \\
        \includegraphics[width=0.31\linewidth]{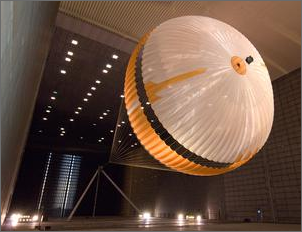} & \includegraphics[width=0.31\linewidth]{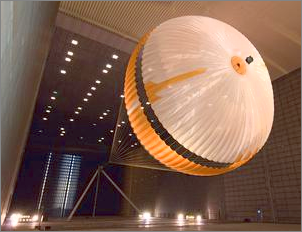} & \includegraphics[width=0.31\linewidth]{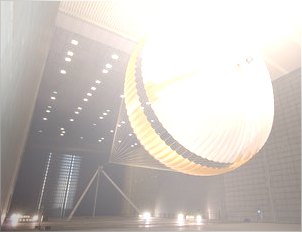} \\[4pt]
        $\mathbf{T0.6}$ & $\mathbf{T0.8}$ & $\mathbf{T1.0}$ \\
        \includegraphics[width=0.31\linewidth]{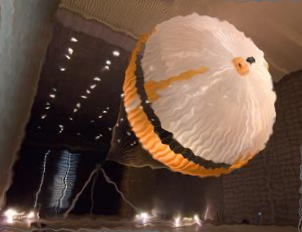} & \includegraphics[width=0.31\linewidth]{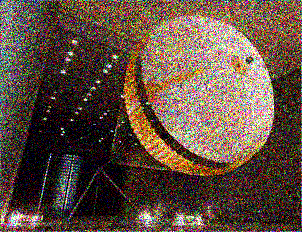} & \includegraphics[width=0.31\linewidth]{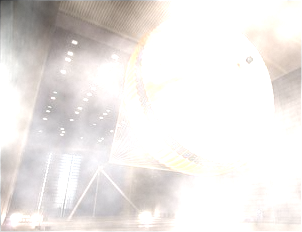} \\
    \end{tabular}
    \caption{Visual degradation severity ladder from Clean to $\mathrm{T1.0}$ for \textbf{M3CoT} (top) and \textbf{ScienceQA} (bottom).}
    \label{fig:corruption_examples_1}
\end{figure}

\begin{figure}[htbp]
    \centering
    \setlength{\tabcolsep}{3pt}
    \renewcommand{\arraystretch}{0.5}
    \begin{tabular}{ccc}
        \multicolumn{3}{c}{\textbf{MathVerse Degradation Ladder}} \\[2pt]
        \textbf{Clean} & $\mathbf{T0.2}$ & $\mathbf{T0.4}$ \\
        \includegraphics[width=0.31\linewidth]{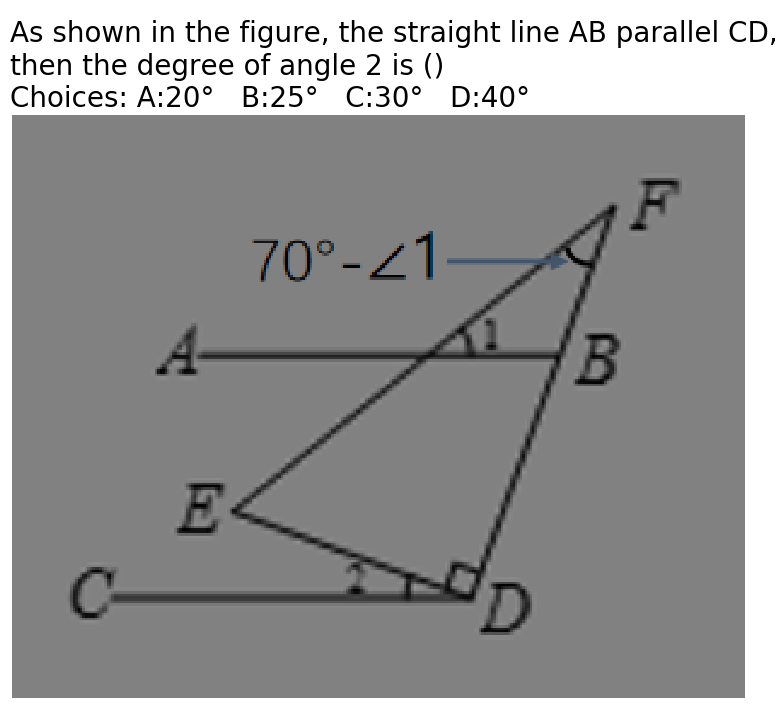} & \includegraphics[width=0.31\linewidth]{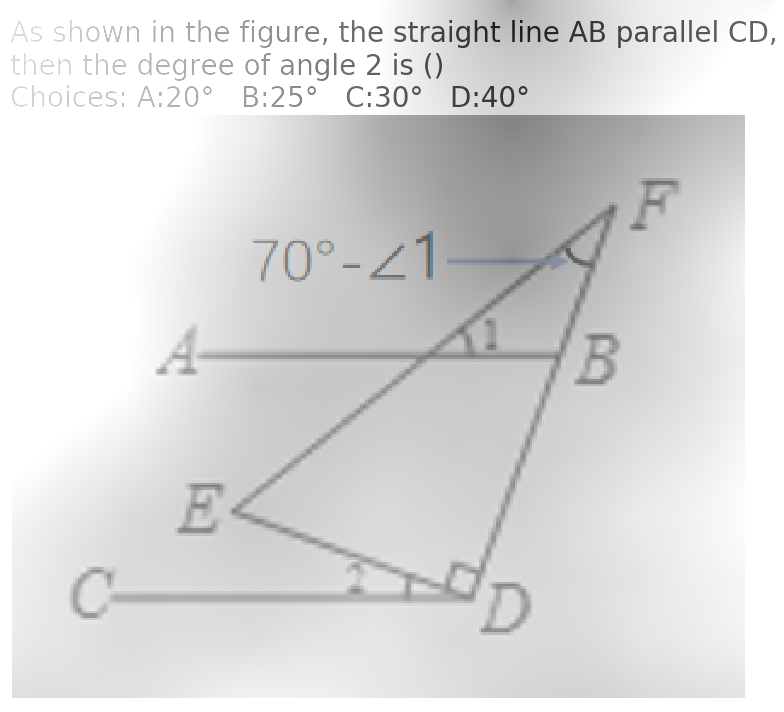} & \includegraphics[width=0.31\linewidth]{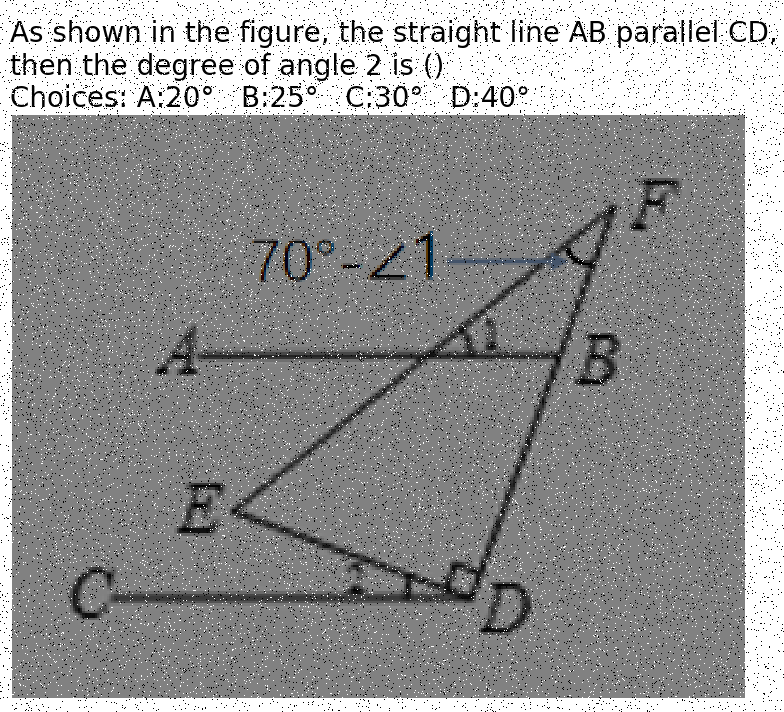} \\[4pt]
        $\mathbf{T0.6}$ & $\mathbf{T0.8}$ & $\mathbf{T1.0}$ \\
        \includegraphics[width=0.31\linewidth]{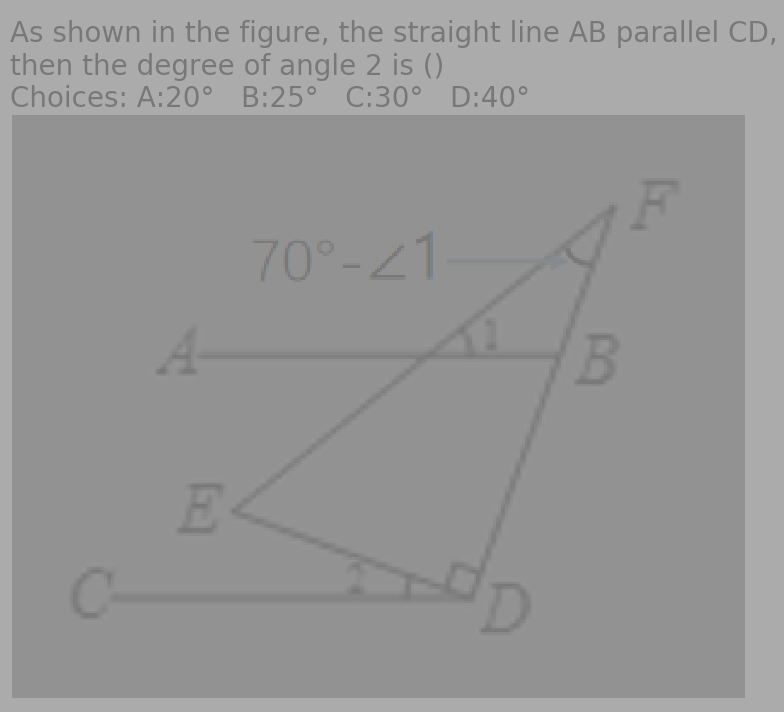} & \includegraphics[width=0.31\linewidth]{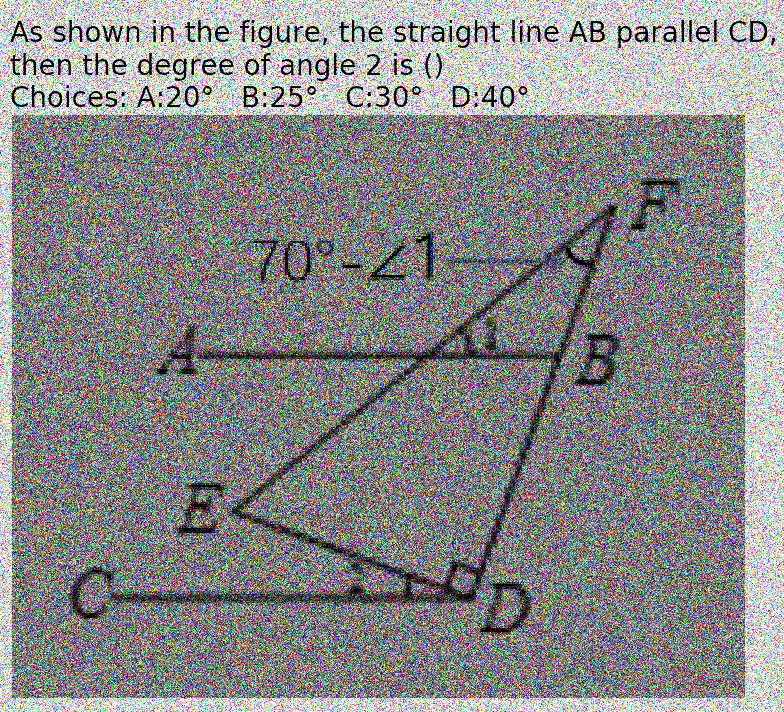} & \includegraphics[width=0.31\linewidth]{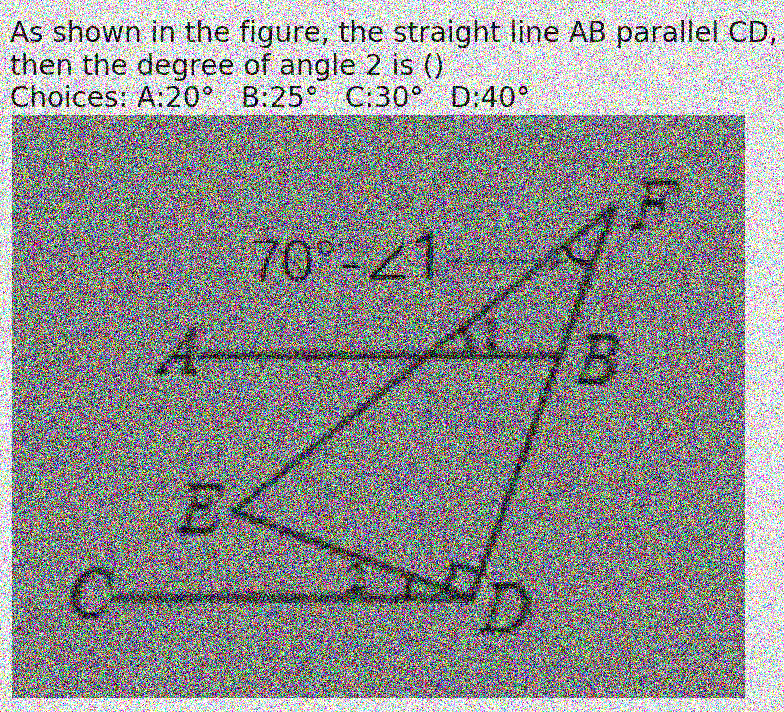} \\[8pt]
        \midrule
        \\[-2pt]
        \multicolumn{3}{c}{\textbf{MMMU Degradation Ladder}} \\[2pt]
        \textbf{Clean} & $\mathbf{T0.2}$ & $\mathbf{T0.4}$ \\
        \includegraphics[width=0.31\linewidth]{figs/image_clean_4_mmmu.jpg} & \includegraphics[width=0.31\linewidth]{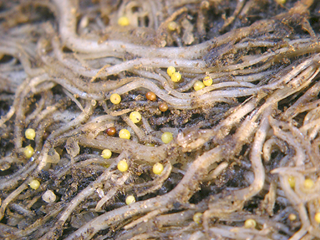} & \includegraphics[width=0.31\linewidth]{figs/image_T0.4_4_mmmu.jpg} \\[4pt]
        $\mathbf{T0.6}$ & $\mathbf{T0.8}$ & $\mathbf{T1.0}$ \\
        \includegraphics[width=0.31\linewidth]{figs/image_T0.6_4_mmmu.jpg} & \includegraphics[width=0.31\linewidth]{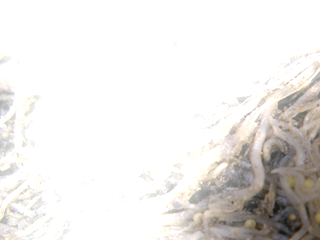} & \includegraphics[width=0.31\linewidth]{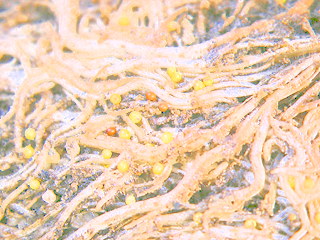} \\
    \end{tabular}
    \caption{Visual degradation severity ladder from Clean to $\mathrm{T1.0}$ for \textbf{MathVerse} (top) and \textbf{MMMU} (bottom).}
    \label{fig:corruption_examples_2}
\end{figure}

\begin{figure}[htbp]
    \centering
    \setlength{\tabcolsep}{3pt}
    \renewcommand{\arraystretch}{0.5}
    \begin{tabular}{ccc}
        \multicolumn{3}{c}{\textbf{We-Math Degradation Ladder}} \\[2pt]
        \textbf{Clean} & $\mathbf{T0.2}$ & $\mathbf{T0.4}$ \\
        \includegraphics[width=0.31\linewidth]{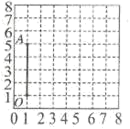} & \includegraphics[width=0.31\linewidth]{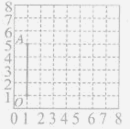} & \includegraphics[width=0.31\linewidth]{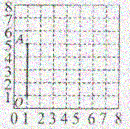} \\[4pt]
        $\mathbf{T0.6}$ & $\mathbf{T0.8}$ & $\mathbf{T1.0}$ \\
        \includegraphics[width=0.31\linewidth]{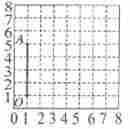} & \includegraphics[width=0.31\linewidth]{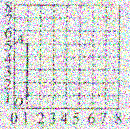} & \includegraphics[width=0.31\linewidth]{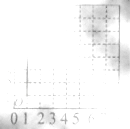} \\[8pt]
        \midrule
        \\[-2pt]
        \multicolumn{3}{c}{\textbf{MathVision Degradation Ladder}} \\[2pt]
        \textbf{Clean} & $\mathbf{T0.2}$ & $\mathbf{T0.4}$ \\
        \includegraphics[width=0.31\linewidth]{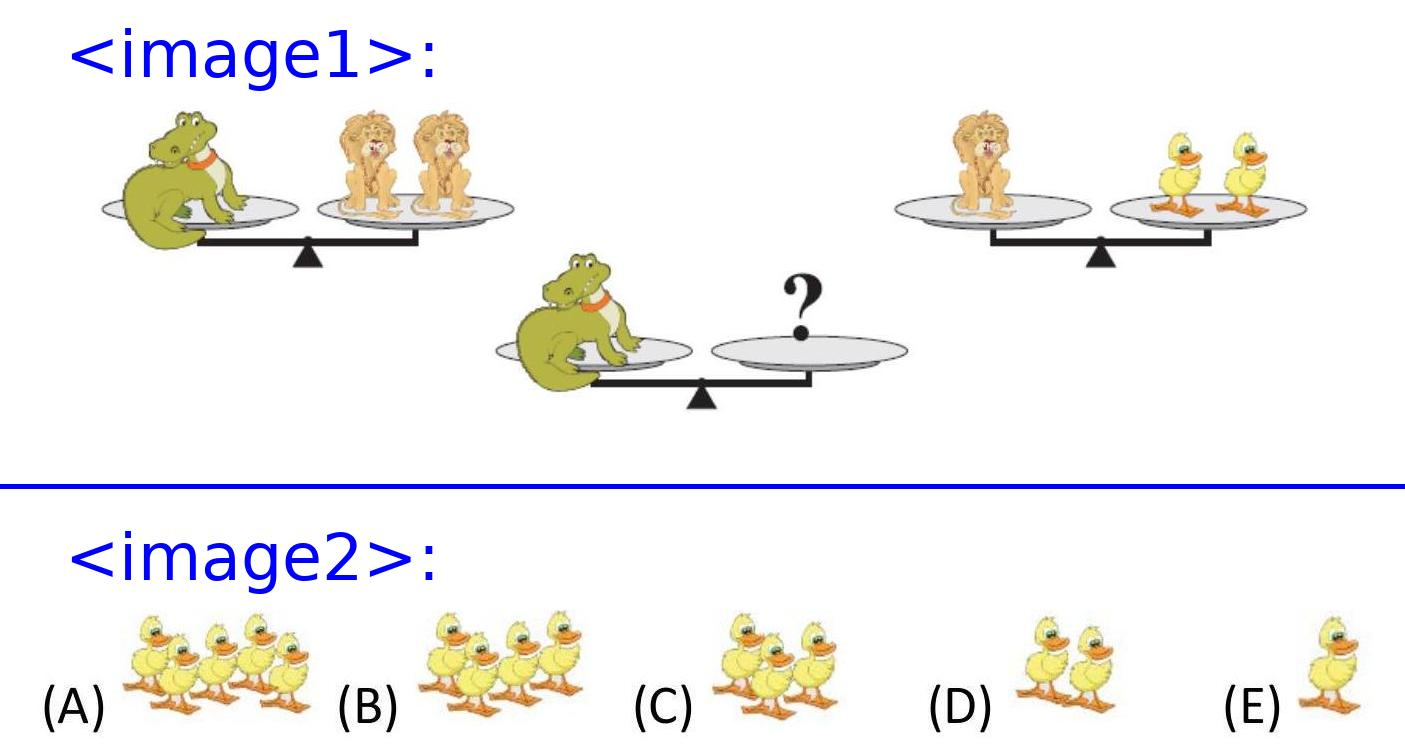} & \includegraphics[width=0.31\linewidth]{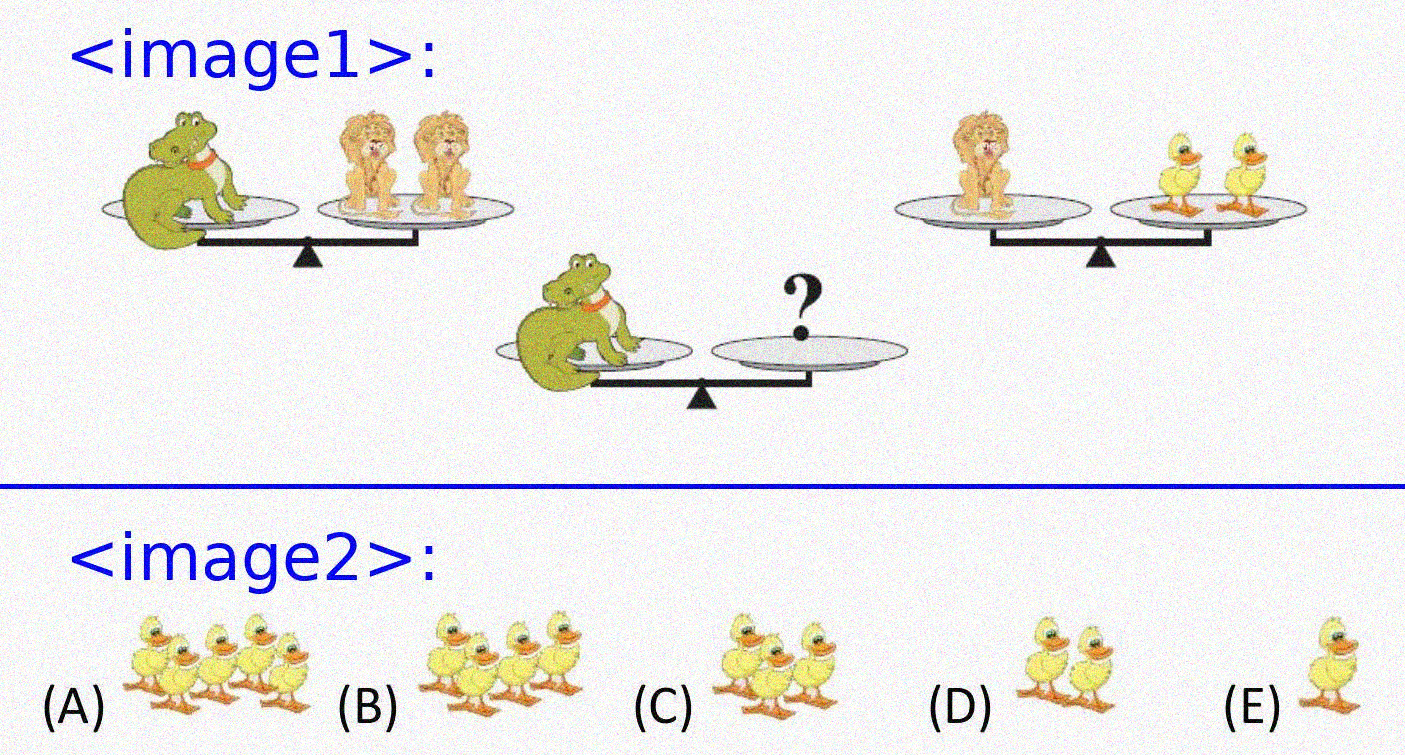} & \includegraphics[width=0.31\linewidth]{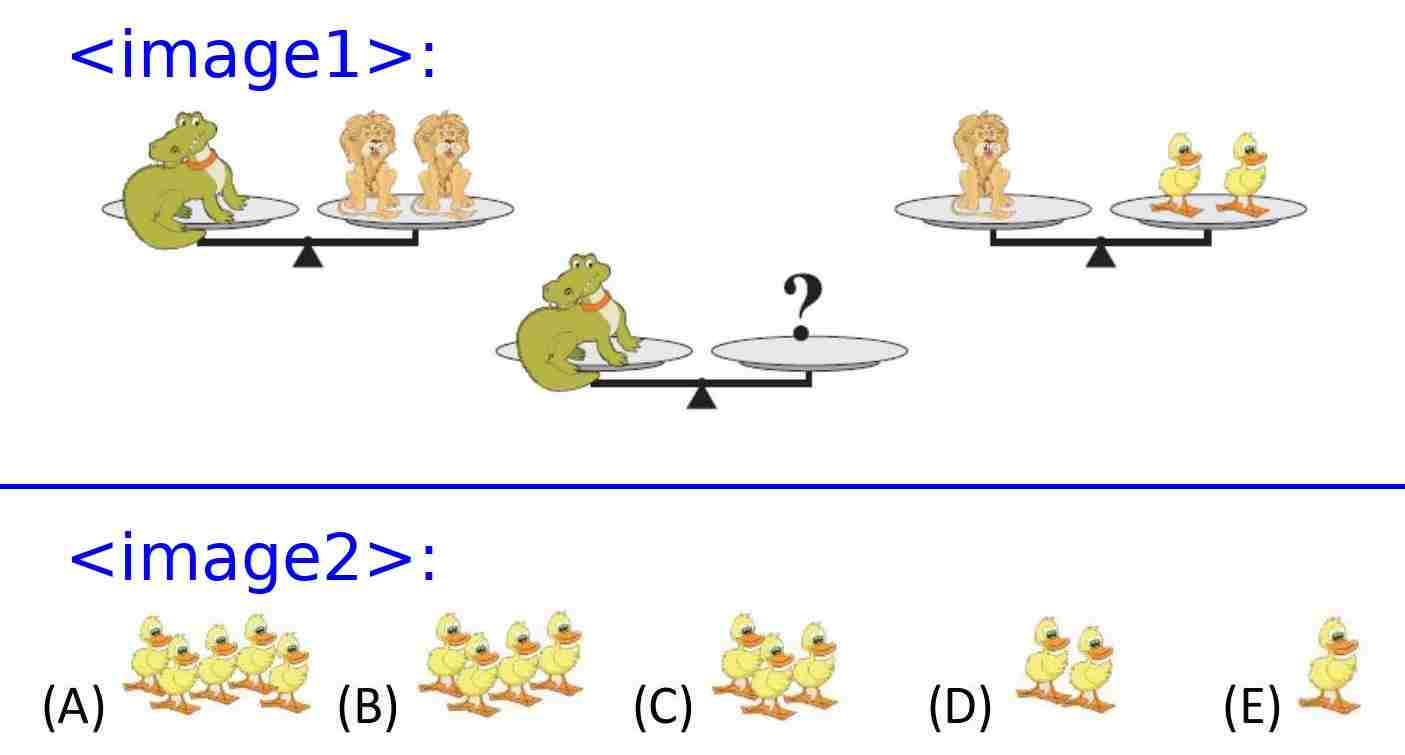} \\[4pt]
        $\mathbf{T0.6}$ & $\mathbf{T0.8}$ & $\mathbf{T1.0}$ \\
        \includegraphics[width=0.31\linewidth]{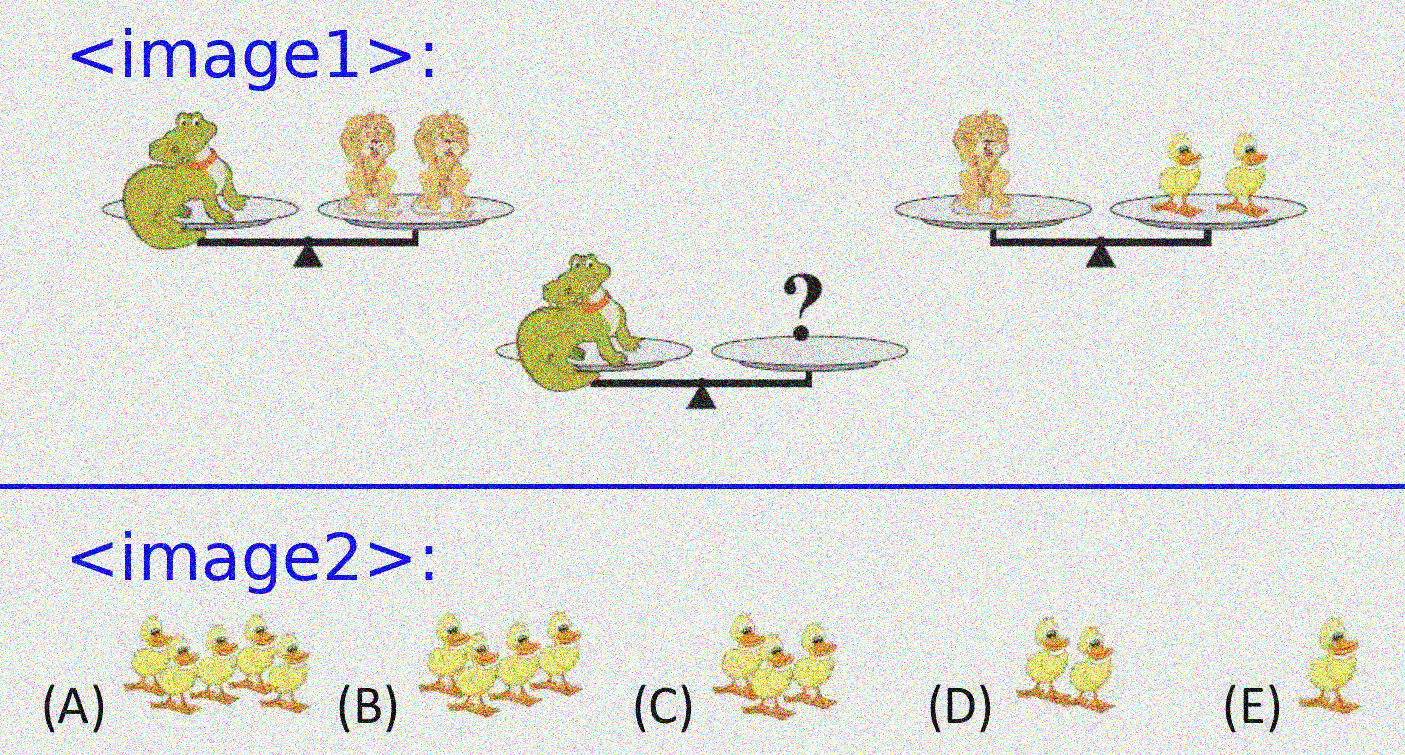} & \includegraphics[width=0.31\linewidth]{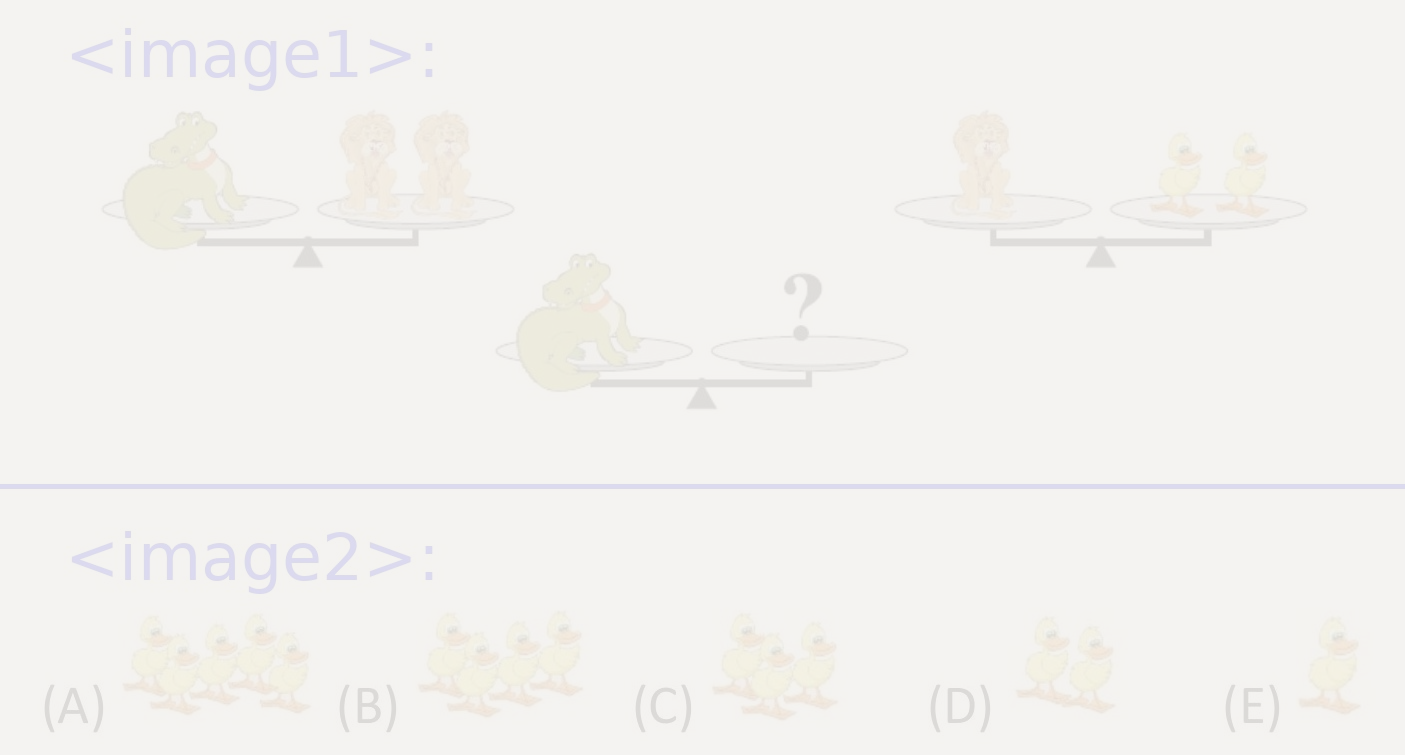} & \includegraphics[width=0.31\linewidth]{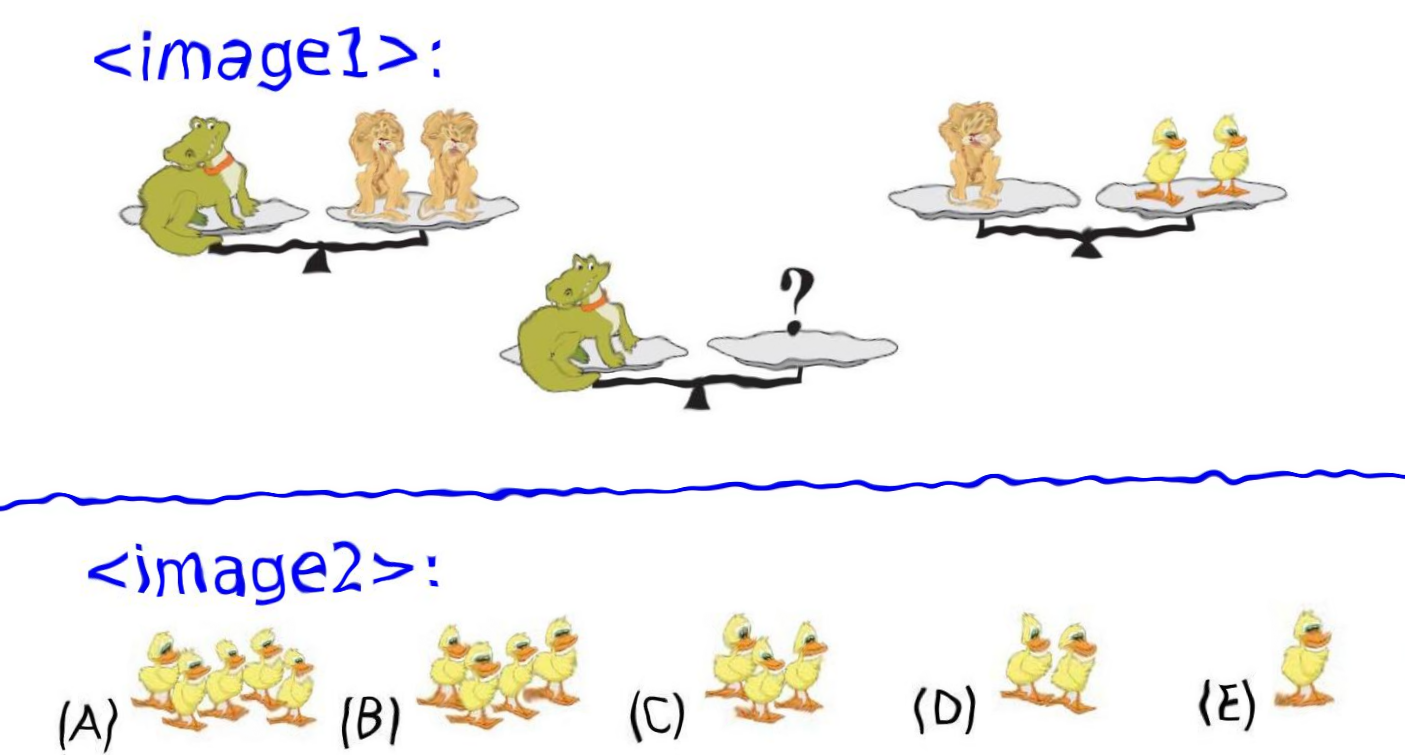} \\
    \end{tabular}
    \caption{Visual degradation severity ladder from Clean to $\mathrm{T1.0}$ for \textbf{We-Math} (top) and \textbf{MathVision} (bottom).}
    \label{fig:corruption_examples_3}
\end{figure}

Figures~\ref{fig:corruption_examples_1}, \ref{fig:corruption_examples_2}, and \ref{fig:corruption_examples_3} demonstrate the effects of increasing corruption severity on samples selected from each dataset. We pair clean base images with their respective $\mathrm{T0.2}$ to $\mathrm{T1.0}$ augmented views to form the 36 evaluation subsets.

\section{Evaluation Metrics}
\label{sec:eval_metrics}
We report three primary summary metrics here. Accuracy measures task success and ignores confidence:
\begin{equation}
\label{eq:acc}
\mathrm{Acc}=\frac{1}{N}\sum_{i=1}^{N} a_i.
\end{equation}
The Brier score measures the overall quality of probabilistic confidence by penalizing both overconfident errors and underconfident correct predictions:
\begin{equation}
\label{eq:brier}
\mathrm{Brier}=\frac{1}{N}\sum_{i=1}^{N}(c_i-a_i)^2.
\end{equation}
Expected calibration error (ECE) measures the average gap between confidence and empirical correctness after binning predictions by confidence:
\begin{equation}
\label{eq:ece}
\mathrm{ECE}=\sum_{b=1}^{B}\frac{|\mathcal{I}_b|}{N}
\left|\frac{1}{|\mathcal{I}_b|}\sum_{i \in \mathcal{I}_b} a_i -
\frac{1}{|\mathcal{I}_b|}\sum_{i \in \mathcal{I}_b} c_i\right|.
\end{equation}
Here $B$ is the number of confidence bins and $\mathcal{I}_b=\{i: c_i \in ((b-1)/B, b/B]\}$ is the index set of predictions whose reported confidence falls into bin $b$. The two averages inside the absolute value are therefore the empirical accuracy and the mean reported confidence within that bin. ECE is small only when each confidence range carries the right empirical success rate. In our setting, $\mathrm{Acc}$ is higher-is-better, whereas Brier and ECE are lower-is-better, so the three metrics separate answer quality from confidence quality under both clean and corrupted inputs.

\section{Compute Resources}
\label{app:compute}
All experiments were conducted on a cluster of 8 NVIDIA Blackwell-series GPUs.
Each training run (one backbone $\times$ one configuration) took approximately 9 hours.
Over the course of this project we conducted approximately 40 experimental runs in total, including exploratory runs and ablation studies, amounting to roughly $40 \times 9 \times 8 = 2{,}880$ single-GPU hours in aggregate.

\section{Use of LLMs}
The drafting of this manuscript was enhanced through the use of a large language model, assisting in grammatical refinement.

\section{Broader Impacts}
The potential negative social impacts of our method are minor, and align with those typically associated with general LLM reasoning technologies. We emphasise the importance of adhering to the principles of fair and safe deployment of LLM systems, including careful consideration of the training data, transparency about model capabilities and limitations, and ongoing monitoring for unintended consequences.

\end{document}